\documentclass[11pt]{article}

\usepackage[preprint]{acl}
\usepackage{hyperref}
\usepackage{xurl}
\usepackage{array}
\usepackage{times}
\usepackage{latexsym}
\usepackage{xcolor}
\usepackage{longtable}
\usepackage{dblfloatfix}
\usepackage{caption}
\usepackage[T1]{fontenc}
\usepackage[utf8]{inputenc}

\usepackage{microtype}

\usepackage{inconsolata}

\usepackage{graphicx}

\usepackage{amsmath}
\usepackage{amssymb}
\usepackage{mathtools}
\usepackage{amsthm}
\usepackage{bbding}

\usepackage{makecell}  % for better cell formatting
\usepackage{adjustbox}

\usepackage{subcaption}
\usepackage{tabularx}
\usepackage{dsfont}
\usepackage{multirow}
\usepackage{wrapfig}

\usepackage{amssymb}
\usepackage{multirow}
\usepackage{color}
\usepackage[capitalize,noabbrev]{cleveref}
\usepackage{lipsum,booktabs}
\usepackage{natbib}

\usepackage{multicol}

\usepackage{caption}
\usepackage{subcaption}
\usepackage{enumitem}

\usepackage{upgreek}

\usepackage{float}

\usepackage{array}

\title{{Revisiting the Effectiveness of LLM Pruning for Test-Time Scaling}}

\author{
\textbf{Ocean Monjur, Shahriar Kabir Nahin, Anshuman Chhabra} \\
  Bellini College of AI, Cybersecurity, and Computing \\
  University of South Florida \\
  Tampa, FL, USA \\
  \texttt{\{omonjur,shahriarkabir,anshumanc\}@usf.edu}\\}

\begin{document}
\maketitle
\begin{abstract}
Large Language Models (LLMs) now exhibit remarkable reasoning capabilities through test-time compute scaling (TTS), with impressive performance across math and coding benchmarks. In parallel, research in model compression has developed \textit{pruning methods} that seek to remove redundant/detrimental parameters without sacrificing task performance. The intersection of these two research advancements lays the foundation for our work.
Specific to reasoning LLMs, prior work has shown that \textit{structured pruning} (methods which remove entire set of layer blocks), significantly degrades TTS reasoning performance. However, in this work, we revisit this assumption and investigate whether \textit{unstructured pruning} (methods that carefully remove only certain redundant/detrimental weights) exhibits similar limitations. 
Surprisingly, our extensive experiments across four reasoning benchmarks on two reasoning LLMs: s1.1-7B and Qwen3-8B, consistently show that unstructured pruning augments TTS performance compared to structured pruning, and at times can even outperform the unpruned full-weight LLMs. Furthermore, we also empirically study the impact of different layer-wise sparsity allocation strategies, which are an important parametric choice for instantiating these unstructured methods. These findings challenge the conventional notion that pruning always reduces TTS performance and in fact, suggest that carefully undertaken pruning can retain TTS effectiveness.

\end{abstract}

\section{Introduction}
\looseness-1 The advent of Large Language Models (LLMs) \cite{NEURIPS2020_1457c0d6} has fundamentally impacted various sectors of society, such as software engineering \cite{khati2025mapping}, education \cite{wang2026large}, and healthcare \cite{LIN2025100868}, among others. Moreover, LLMs attain improvements in capabilities as a function of scale (in terms of both data and parameter size) \cite{kaplan2020scalinglawsneurallanguage}, thus necessitating ever-increasing computational resource requirements for model storage, training, and inference \cite{bai2024efficiencysystematicsurveyresourceefficient}. 

% To mitigate these challenges and reduce the computational costs associated with model {development}, the research community has {proposed}
Motivated by these large model sizes, the research community has also proposed several approaches for pruning large neural networks such as LLMs \cite{lecun1989optimal,hassibi1993optimal,sun2024a,lu2024alphapruning}. In essence, state-of-the-art LLM pruning methods seek to eliminate redundant or less influential model parameters, with minimal to little impact on performance \cite{askari2025layerif}. Furthermore, LLM pruning can generally be categorized into two types: \textit{structured pruning} \cite{men2025shortgpt}, which removes entire model layer blocks, and \textit{unstructured pruning} \cite{sun2024a}, which removes individual weights inside layer blocks while following predefined sparsity rates per layer.

In complementary work aimed at further {improving} LLM performance, recent work has found that allocating additional computation during inference significantly increases downstream task performance on a myriad of complex reasoning tasks (e.g., math and coding problems) \cite{muennighoff2025s1}. This inference paradigm, commonly referred to as test-time scaling (TTS), increases model performance by allowing it to generate additional tokens sequentially and decompose problems into reasoning traces \cite{wei2022chain}.% or sample from several candidate solutions using efficient search algorithms (e.g., Monte Carlo Tree Search) \cite{wei2022chain,wang2023selfconsistency}.

Our work is primarily concerned with the \textit{intersection} of both these adjacent but complementary research directions (i.e., LLM pruning and TTS). In particular, recent work \cite{wang2025fewer} has explored the effect of \textit{structured} LLM pruning in TTS and has found that while pruned models retain performance in short or shallow reasoning tasks, significant performance degradation is observed in tasks that require longer reasoning chains. Their findings suggest that structured LLM pruning significantly impairs the model's ability to handle long and complex multi-step reasoning tasks.

\begin{figure*}[!t]
    \centering
    \includegraphics[width=.95\textwidth]{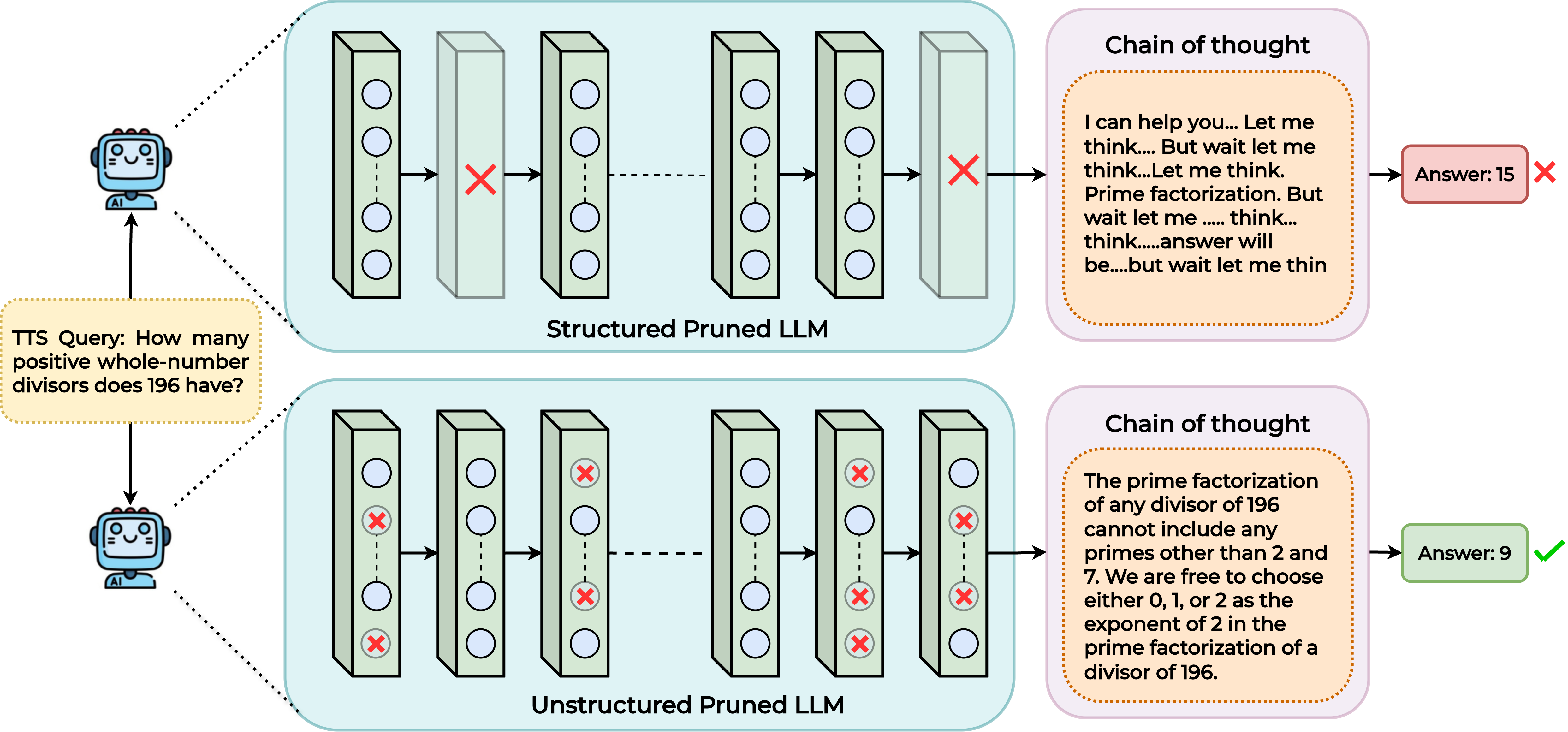}\vspace{-2mm}
    \caption{Overview of \textit{structured} and \textit{unstructured} pruning for LLMs and their impact on test-time scaling (TTS) reasoning performance. As identified in prior work, removing entire layer blocks via \textit{structured} pruning makes LLMs more susceptible to producing incoherent chains of thought, ultimately resulting in incorrect answers. However, as our findings show, this is not the case for \textit{unstructured} pruning, where TTS performance can be augmented significantly via targeted weight removal.}\vspace{-2mm}
    \label{fig:intro}
\end{figure*}

\looseness-1 We move beyond the structured pruning case and demonstrate that while structured pruning degrades performance for TTS, this trend does not hold for recently proposed \textit{unstructured} LLM pruning methods, such as Magnitude \cite{NIPS2015_ae0eb3ee} and Wanda \cite{sun2024a} pruning. Interestingly, we find that unstructured pruning strategies can match \textit{unpruned} LLMs' performance in most cases, and at times, even exceed their performance on some benchmarks, thus demonstrating their potential in augmenting LLMs' reasoning performance despite possessing fewer parameters. Through our work, we seek to galvanize future research that can fully disentangle the role of unstructured pruning for performance improvements in next-generation LLMs. %utilize unstructured pruning methods as a strategy for operating even larger next-generation LLMs in a tractable manner. 

In sum, we make the following contributions:

\begin{itemize}[nosep]
    \item Through extensive experiments on the MATH500 \cite{hendrycks2021measuring,lightman2024lets}, AIME24 \cite{AIME}, AMC23 \cite{liao-etal-2025-enhancing}, and GPQA-Diamond \cite{rein2024gpqa} datasets using the s1.1-7B \cite{muennighoff2025s1} and Qwen3-8B \cite{yang2025qwen3technicalreport} LLMs we show that, contrary to prior assumptions, not all pruning methods result in detrimental TTS performance.

    \item Our findings confirm that although \citet{wang2025fewer} correctly identified that structured pruning degrades TTS performance, their findings do not hold for \textbf{\textit{unstructured pruning}} strategies, which at times can even exceed the performance of unpruned models.
    
    \item We empirically investigate different layer-wise sparsity allocation strategies in unstructured pruning to assess their impact on TTS performance and derive novel insights with regards to their usefulness.

\end{itemize}

\section{Related Works}
\looseness-1\textbf{LLM Pruning.} Neural network pruning serves as a means for reducing model complexity while preserving predictive performance \cite{lecun1989optimal,hassibi1993optimal}. These approaches have also shown significant promise in LLMs. Unstructured pruning approaches, including classical magnitude-based \cite{NIPS2015_ae0eb3ee} pruning as well as recent methods such as Wanda \cite{sun2024a}, and SparseGPT \cite{pmlr-v202-frantar23a},  operate by selectively masking individual weights based on some predefined criteria. Magnitude pruning prunes weights that have the lowest weight magnitude, while Wanda masks based on the input activations multiplied by the weight magnitudes. SparseGPT, an optimization-based method, prunes sequentially by minimizing the layer-wise output reconstruction error. Structured pruning methods such as ShortGPT \cite{men2025shortgpt}, LLM-Pruner \cite{NEURIPS2023_44956951}, and SliceGPT \cite{ashkboos2024slicegpt} work by removing entire layers based on some measures of importance.\vspace{2mm}% ShortGPT utilizes block-influence to sort and prune specific layers. While LLM-Pruner is an optimization-based pruning approach.\vspace{2mm}

\looseness-1\noindent \textbf{Test-Time Scaling.} Utilizing additional compute in LLMs during inference via the generation of \textit{thinking tokens}, has shown to yield significant performance improvements on tasks requiring complex reasoning \cite{wei2022chain,muennighoff2025s1,chen2025provable,kimiteam2025kimik15scalingreinforcement}. Several such TTS strategies have been proposed. Search/verification approaches explore multiple candidate solutions through search \citep{cobbe2021trainingverifierssolvemath, lightman2023let, coulom2006efficient,gao2024interpretablecontrastivemontecarlo}, ensembling strategies such as PackLLM \citep{mavromatis2024pack} use perplexity-based weighting for test-time model fusion, iterative refinement methods such as Self-Refine \citep{madaan2023self} undertake continuous revision, and temperature-based methods \citep{xie-etal-2024-calibrating} adjust token generation at inference, among others. In the context of LLM pruning and TTS, \cite{wang2025fewer} study only structured pruning methods (e.g. ShortGPT \cite{men2025shortgpt}) and find that they deteriorate TTS performance. In this work, we extend this analysis to unstructured pruning and explore how different parametric choices impact downstream TTS capabilities.

%, that seek to allocate compute during inference have been proposed including approaches based on longer chain-of-thought generation, self-consistency sampling, and iterative reasoning \cite{zuo2025ttrl,wang2025fewer}. These strategies scale reasoning performance by increasing inference time computation, allowing models to explore multiple reasoning paths or deeper reasoning processes without modifying the underlying model parameters.

%can be divided into sequential and parallel scaling \cite{zuo2025ttrl,wang2025fewer}. Sequential scaling increases inference time computation by allowing the model to generate longer responses, whereas parallel scaling allocates additional computation by producing multiple candidate outputs in parallel.

%\section{Problem Statement}

\section{Preliminaries and Background}

In this section, we introduce preliminaries and provide background on LLM pruning methods and TTS strategies.

\subsection{Test-Time Scaling (TTS)}
\looseness-1TTS enhances model performance by allocating additional compute at test-time, essentially increasing the number of generated tokens at the output to reason more about the input query.
Let $x \in X$ be an input, and let $C_1 < C_2 < C_3 ...< C_k$ denote a sequence of token limits. A model $M$ under gradual increase in token limits $C_j$ produces an output: $y^{C_j} = M(x;C_j)$ where $y^{C_j}$ is the sequence generated by model $M$ up to $C_j$ tokens. Such a scaling of test-time compute has been shown to result in improved performance on complex reasoning tasks.

 % \looseness-1 \noindent\textbf{Parallel TTS.} For a model $M$ in parallel test-time scaling, it generates parallel sequences given input $x \in X$. Let $G(x; M)$ denote the set of possible sequences generated by $M$ under parallel evaluation. A reward model $r: V^* \to \mathbb{R}$ assigns a scalar score to each candidate sequence. $y^\star = T_{\text{par}}(x; M, r, G(x; M))$ where $y^\star$ is the output that is chosen.

\subsection{Large Language Model Pruning} 

\looseness-1 Model pruning works by reducing or removing redundant or less influential parameters from a model while attempting to maintain its original performance. For LLMs, pruning can be categorized as: \textit{Structured Pruning} and \textit{Unstructured Pruning}.

\subsubsection{Structured Pruning}

Structured pruning removes entire layers of an LLM, resulting in a reduced network with a smaller architecture. Let $L\subseteq\{1,..., N\}$ be the set of retained layer indices, selected by some criterion $\phi$ and some threshold $\tau$:
\begin{equation}
L = \{ l: \phi(W^{(l)}) \geq \tau \}
\label{eq:structured}
\end{equation}

Then the structured pruned model $\tilde M_{struct}$ is $\tilde M_{struct} = \tilde W^{(l)}$ where $\tilde W^{(l)} = \{W^{(l)}\}_{l \in L}$.\vspace{1.5mm}

\noindent\textbf{ShortGPT} \cite{men2025shortgpt}: is a structured pruning approach which removes layers according to \textit{Block Influence} (BI) scores where layers with the smallest BI scores are removed, and the BI score of layer $i$ is defined as:\vspace{-3mm}

\begin{equation*}
BI_i = 1 - \mathbb{E}_{X,t} \frac{X_{i,t}^{T}X_{i+1,t}}{||X_{i,t}||_2||X_{i+1,t}||_2}
\end{equation*}

\looseness-1\noindent where $X_i,t$ is the $t^{th}$ row of hidden states of the $i^th$ layer. Therefore, in \cref{eq:structured}, the $BI_i$ scores act as $\phi$ and the number of layers to remove is $\tau$.

\subsubsection{Unstructured Pruning}
Unstructured pruning masks individual weights independently of the network’s architecture, resulting in a sparse model that retains the original structure. Let $S^{(l)} \in \{0,1\}$ be a binary mask applied to the weight matrix $W^{(l)}$ of layer $l$ which can be defined as $\tilde W^{(l)} = S^{(l)} \odot  W^{(l)}$. Here the mask is defined by some threshold $\tau$ and $\phi$ as pruning criterion:
\begin{equation}
S^{(l)}_{ij} = \begin{cases} 1 & \text{if } \phi (W^{(l)}_{ij}) \geq \tau \\ 0 & \text{otherwise} 
\end{cases}
\label{eq:unstructured}
\end{equation}

\noindent Then the unstructured pruned model $\tilde M_{unstruct}$ is, $\tilde M_{unstruct} = \tilde W^{(l)}$.\vspace{1.5mm}

\noindent\textbf{Magnitude Pruning} \cite{NIPS2015_ae0eb3ee}: is an unstructured pruning approach in which a specified fraction of the weights with the smallest magnitudes are masked out. Formally, given a desired sparsity level $s$, the bottom $s\%$ of the weights (by absolute value) are masked, producing a sparse model while preserving the original structure of the network. Clearly in \cref{eq:unstructured} the criterion $\phi$ can simply be defined as: $\phi_{mag} = |W|$.\vspace{1.5mm}

\noindent\textbf{Wanda} \cite{sun2024a}: is an improvement upon magnitude pruning, where the pruning criterion is defined as the elementwise product of the weight magnitudes $W$ and the norms of the input activations $||X||_2$.
From \cref{eq:unstructured} the criterion $\phi$ for Wanda is defined as, $\phi_{wan} = |W| \odot ||X||_2$.\vspace{2mm}

\looseness-1\noindent\textbf{Note on Sparsity Allocation Strategies.} While the default strategy when utilizing the aforementioned unstructured pruning methods is to prune all layers \textit{\textbf{uniformly}}, several layer-specific sparsity allocation methods have also been proposed that prune different layers at different rates (totaling the overall pruning ratio) for improved performance. To this end, alongside uniform pruning, in this work we will experiment with two popular and recently proposed sparsity allocation methods: \textbf{Outlier Weighted Layerwise Sparsity (OWL)} \cite{pmlr-v235-yin24e} and \textbf{LayerIF} \cite{askari2025layerif} to assess the impact on downstream TTS performance. %Furthermore, our research questions and corresponding experiments will also assess the impact of different layer choices (e.g. MLP versus attention layers) for unstructured pruning and how that impacts downstream TTS performance.

\section{Proposed Research Questions}

%\textbf{Research Questions.}
We now define the two fundamental research questions (RQs) we aim to study in our work.

% \noindent 

% \begin{itemize}
%   \item \textbf{Research Question \#1 (RQ1)}:\\
%   \fbox{%
%     \parbox{0.455\textwidth}{\textit{Does \textbf{unstructured pruning} hinder the effectiveness of TTS in LLMs, similar to structured pruning (as shown in recent work by \citet{wang2025fewer})? If not, can unstructured pruning preserve or improve TTS performance further?}}}

%   \item \textbf{(RQ2)} \textit{For unstructured pruning, which \textbf{layer sparsity allocation} strategies (such as uniform pruning across layers, influence-based allocation methods such as LayerIF \cite{askari2025layerif}, etc.) and \textbf{pruning layers} (attention heads or feedforward MLP layers) lead to improved TTS performance?}

% \end{itemize}

\begin{itemize}[nosep]
  \item \textbf{(RQ1)}
  {\textit{Does \textbf{unstructured pruning} hinder the effectiveness of TTS in LLMs, similar to structured pruning (as shown in recent work by \citet{wang2025fewer})? If not, can unstructured pruning preserve or further improve TTS performance?}}

  \item \textbf{(RQ2)} \textit{For unstructured pruning, which \textbf{layer sparsity allocation} strategies (uniform pruning across layers, influence-based allocation methods such as LayerIF \cite{askari2025layerif}, and outlier methods such as OWL \cite{pmlr-v235-yin24e}) lead to improved TTS performance?}
\end{itemize}

\looseness-1 These research questions establish the scope and objectives of our work, and inform the methodology and experimental evaluations in the subsequent sections.

%write down the RQs in English as we had discussed and then also explain them mathematically

% Pruning is widely regarded as an effective technique to reduce the computational load of large language models while maintaining their performance \cite{lu2024alphapruning,sun2024a}. Prior work on pruning in the context of test-time scaling has shown that pruning can severely degrade model performance \cite{wang2025fewer}. While the application of structured pruning has been explored in test-time scaling, the performance of unstructured pruning has been largely unexplored in the previous literature.

% It is evident that structured pruning, which removes entire layers or components, significantly impairs the model's ability to perform long-chain reasoning. Unstructured pruning does not remove whole layers from a model, which enables LLMs to retain their performance on these problems. 

%In this paper, we explore the viewpoint that even if structured pruning is harmful for LLMs in test-time scaling, unstructured pruning can be beneficial. Mathematically we can define our research question as,

%\begin{equation}
%    P_{TTS}(\tilde M_{struct}) < P_{TTS}( M_{unpruned})
%\end{equation}
 
%\begin{equation}
%    P_{TTS}( M_{unpruned}) \leq P_{TTS}(\tilde M_{unstruct})
%\end{equation}

%Where $P_{TTS}$ indicates performance on a test-time scaling task, $M_{unpruned}$ is the unpruned model, $\tilde M_{struct}$ is the structured pruned model and $\tilde M_{unstruct}$ is the unstructured pruned model.

\section{Experiments and Results}

\looseness-1\textbf{Datasets.}
To empirically evaluate our RQs, we conducted experiments on four widely used reasoning benchmarks that span advanced mathematical and scientific domains: MATH500 \cite{hendrycks2021measuring,lightman2024lets}, AIME24 \cite{AIME}, AMC23 \cite{liao-etal-2025-enhancing}, and GPQA-Diamond \cite{rein2024gpqa}. These benchmarks contain challenging reasoning tasks that require multi-step inference, making them ideal candidates for studying the impact of pruning under TTS.

\begin{figure*}[!t]
    \centering
    \includegraphics[width=0.98\textwidth]{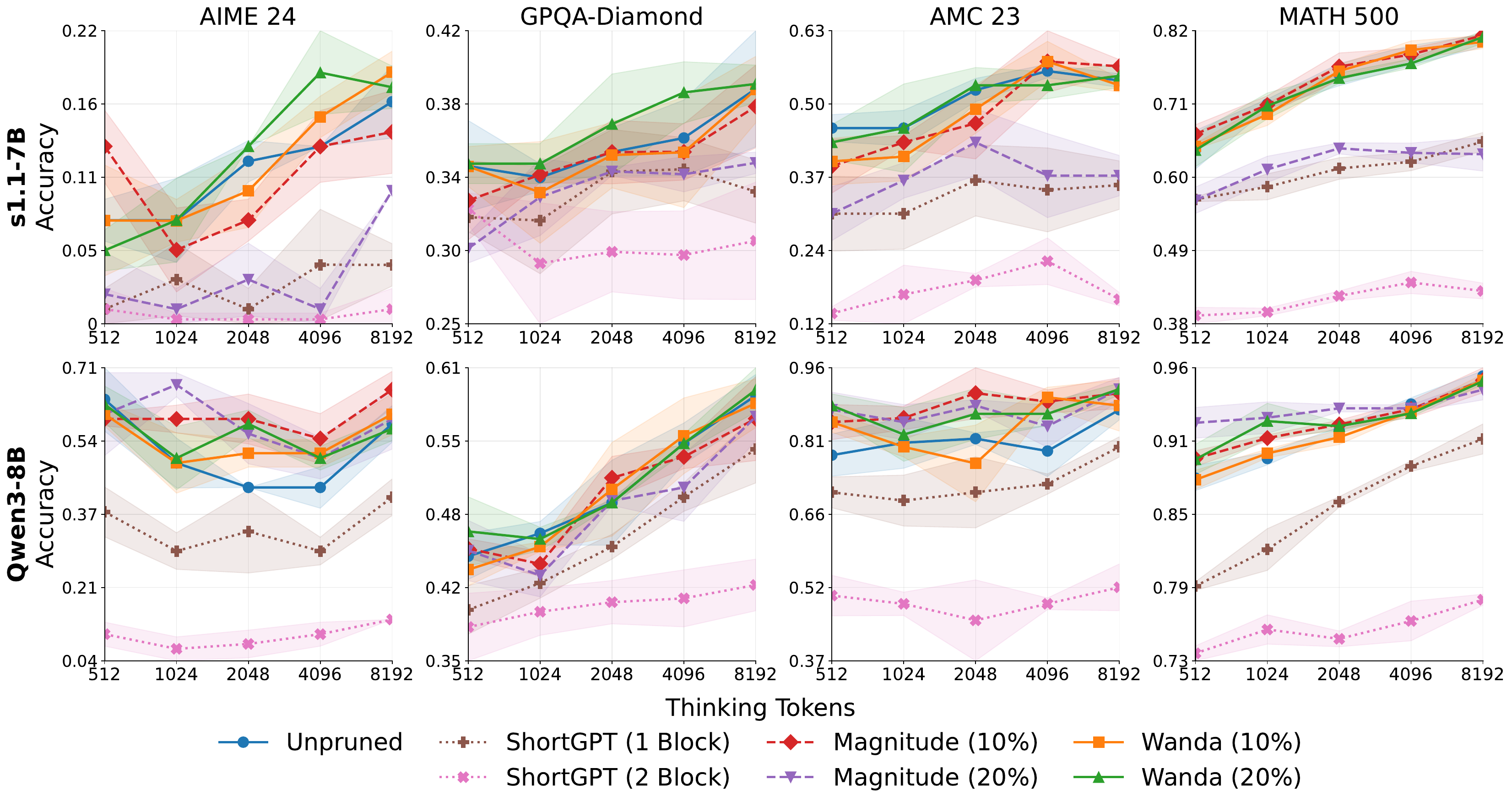}\vspace{-3mm}
    \caption{Comparing \textit{structured} (\textit{ShortGPT}) and \textit{unstructured} \textit{(Magnitude, Wanda)} pruning methods on four long-chain reasoning datasets. Unstructured pruning is employed \textit{uniformly} at both 10\% and 20\% sparsity rates, while structured pruning removes 1 and 2 layer blocks. It is evident that unstructured pruning retains or surpasses unpruned LLM performance, whereas structured pruning leads to substantial degradation.}\vspace{-2mm}
    \label{fig:RQ1}
\end{figure*}

\noindent \textbf{LLMs.} To study the effect of pruning under TTS, we conduct experiments using open-source models with strong reasoning capabilities, such as  s1.1-7B \cite{muennighoff2025s1} and Qwen3-8B \cite{yang2025qwen3technicalreport}. The s1.1-7B model is trained on the high-quality s1k \cite{muennighoff2025s1} reasoning dataset, which introduces budget forcing during inference for improved TTS. Qwen3-8B LLM is also a reasoning LLM with state-of-the-art multi-step reasoning capabilities. Both models have previously been used to compare structured pruning against the base unpruned model in TTS \cite{wang2025fewer}, which makes them well-suited for our analysis as well. Moreover, owing to space constraints, we also provide additional results for Llama-3.1-8B in Appendix \ref{appendix:Additional Results}.  %\textcolor{red}{Additional results on Llama-3.1-8B (Deepseek-R1-distil) is provided in Appendix \ref{appendix:Additional Results}}.

\noindent \textbf{Methodology and Protocol.}
Our experiments follow a similar methodology to \citet{wang2025fewer}. In our experiments, we evaluate each of the model configurations (unpruned, different structured and unstructured pruning methods) under sequential TTS at five different thinking token limits: 512, 1024, 2048, 4096, 8192. All results are obtained over 3 runs with random seeds. Furthermore, we conduct experiments on standard sparsity ratios (10\% and 20\%) for unstructured pruning approaches (Wanda and Magnitude) as per prior work \cite{zhang2024structured,zhu2026high}. Note that for structured pruning (e.g. ShortGPT) parameters are pruned as entire layer blocks, so vary from 2.7\%-7\% in terms of total parameter counts.\footnote{While these structured pruning rates of 2.7\% (1 layer) and 5.7\% (2 layers) for Qwen3-8B, and 3.5\% (1 layer) and 7\% (2 layers) for s1.1-7B, are non-standard for unstructured pruning, we also provide results for these pruning rates in Appendix \ref{appendix:Same_Sparsity} and obtain similar trends as our main experiments.} Subsequently, we will investigate different sparsity allocation strategies such as Uniform, OWL, and LayerIF and their impact on TTS.

\looseness-1As prescribed in \citet{pmlr-v235-yin24e}, for OWL, we set $M = 7$, where $M$ is defined as the value that determines the cutoff point for identifying outlier weights. For LayerIF \cite{askari2025layerif}, we allocate sparsity to both attention and MLP layers. Moreover, for computational efficiency, we assume the Hessian to be the identity matrix as in prior work on influence analysis \cite{NEURIPS2020_e6385d39, chhabra2025oga, vitel2025first}. Due to space constraints, we provide additional implementation details in Appendix~\ref{appendix:Implementation}.

% Finally, we also investigate how pruning specific parts, such as attention heads or feedforward MLPs in isolation, affects the overall performance of LLMs in test-time scaling across different reasoning tasks. Our comprehensive comparisons help identify which strategies best preserve reasoning capabilities under varying levels of sparsity on test-time scaling.

\subsection{(RQ1) Impact of Unstructured Pruning on TTS Performance}
\looseness-1 \cref{fig:RQ1} presents a comparison of structured and unstructured pruning approaches for TTS, highlighting their relative impact on performance. The results for structured pruning (ShortGPT) are largely similar to those obtained by \cite{wang2025fewer}, showing substantial performance degradation across all four datasets for both reasoning models compared to the unpruned LLM baselines. The results obtained for unstructured pruning approaches (Wanda and Magnitude) however are substantially different. Not only do these methods outperform their structured pruning counterparts, they also consistently match or even exceed the performance of the unpruned models across all datasets and reasoning models.

\looseness-1 Focusing on the s1.1-7B model, we observe that unstructured pruning using Wanda-20\% yields substantially better performance across most thinking-token budgets compared to the unpruned variant on AIME24 and GPQA-Diamond. For AMC 23 and Math500, Wanda-20\%, Wanda-10\%, and Magnitude-10\% closely align or slightly outperform the unpruned baseline.

The Qwen3-8B model generally achieves substantially higher performance than s1.1-7B on the selected reasoning datasets, making comparisons across pruning strategies more robust. For Qwen3-8B, we can observe that the unstructured pruning approaches: Wanda-20\%, Magnitude-10\%, and Magnitude-20\%, show significant improvements over the base unpruned model on AIME24 and AMC 23, while results on GPQA-Diamond and MATH500 show equivalent or slightly increased performance. While Magnitude-20\% showed weaker performance compared to other unstructured pruning approaches on the other reasoning model s1.1-7B, for Qwen3-8B, its results are comparable with other unstructured pruning methods.

\begin{figure*}[!t]
    \centering
    \includegraphics[width=0.98\textwidth]{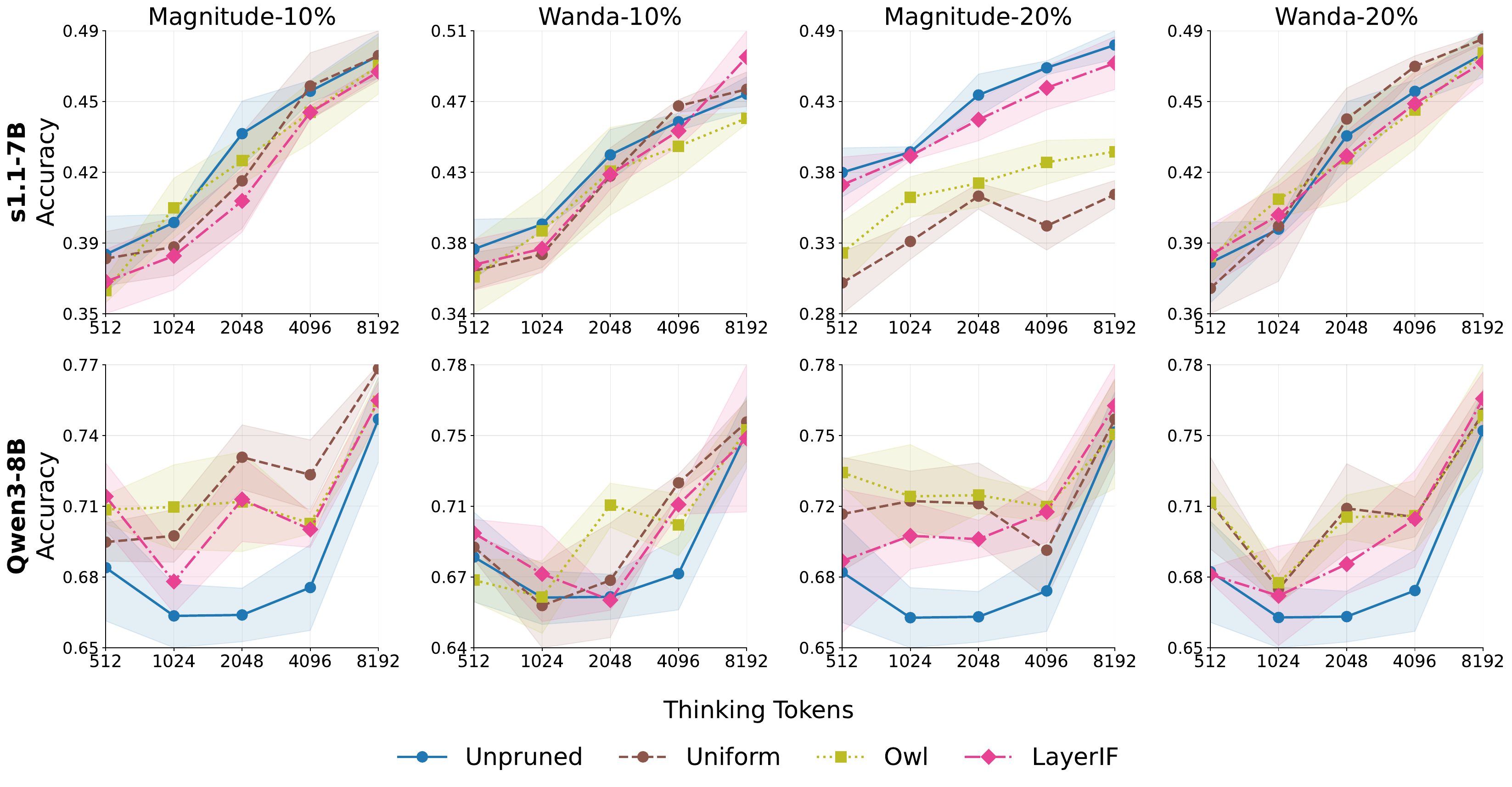}\vspace{-3mm}
\caption{Comparing different \textit{layer-wise sparsity allocation} strategies \textit{(Uniform, OWL, and LayerIF)} with global sparsity rates of 10\% and 20\%. Performance is averaged across AIME24, GPQA-Diamond, AMC23, and MATH500 benchmarks while varying thinking tokens from 512 to 8192.}\vspace{-2mm}
    \label{fig:RQ2}
\end{figure*}

\looseness-1Across both models, we observe a consistent trend in which unstructured pruning methods surpass the unpruned baseline, with performance gains of upto $\approx$10\% for certain token budgets (1024, 2048, and 4096) in Qwen3-8B. This suggests that these methods effectively identify and remove weights detrimental to TTS while preserving those that are beneficial for the same. For GPQA-Diamond, we observe the same pattern: unstructured pruning consistently surpasses the unpruned baseline with Wanda-20\% attaining the best performance for s1.1-7B. A similar trend of unstructured approaches matching or exceeding unpruned models holds for AMC23. We observe that the performance of unstructured pruned models matches that of the unpruned baseline for s1.1-7B, while for Qwen3-8B, they consistently exceed it. The results for MATH500 across both models show that the unstructured approaches closely match the unpruned models. We attribute this behavior to the initial superlative performance of the unpruned model, which leaves little scope for additional improvements. But this result on MATH500 demonstrates that degradation is almost nonexistent, making unstructured pruning an exceptionally strong approach for improving/preserving TTS capabilities.%\vspace{1mm}

\noindent \textbf{Remarks on RQ1 Findings.} Experiments across all four reasoning datasets and both LLMs demonstrate a consistent conclusion: \textit{unstructured pruning} not only outperforms \textit{structured pruning}, but also retains or, on most occasions, outperforms the performance of the unpruned LLMs. These findings provide greater clarity and, in essence, disagree with the conclusions of \citet{wang2025fewer}, who argue that pruning, particularly in the context of TTS, generally leads to performance degradation. While our results confirm their observation that \textit{structured pruning} is detrimental for test-time scaling, clearly this conclusion does not generalize to all pruning methods, especially those that are {\textit{unstructured}} in nature.

%\begin{figure}[t]
%    \centering
%    \includegraphics[width=0.48\textwidth]{images/RQ2.pdf}\vspace{-3mm}
%    \caption{Comparing different \textit{layer-wise sparsity allocation} strategies \textit{(Uniform, Owl, and LayerIF)} with a global sparsity rate of 20\%. Performance is averaged across AIME24, GPQA-Diamond, AMC23, and MATH500 benchmarks while varying thinking tokens from 512 to 8192.}\vspace{-3mm}
%    \label{fig:RQ2}
%\end{figure}

\begin{table*}[!t]
\centering
\small
\caption{Results for the s1.1-7B and Qwen3-8B LLMs obtained across layer-wise sparsity allocation strategies and unstructured pruning approaches with a total/global sparsity rate of 20\%, and maximum thinking tokens = 8192. Top performing configuration for each LLM and dataset in bold.}\vspace{-1.5mm}
\begin{tabular}{lll|cccc}
\toprule
\textbf{Model} & \textbf{Allocation} & \textbf{Pruning} & \textbf{AIME24} & \textbf{GPQA-Diamond} & \textbf{AMC23} & \textbf{MATH500} \\
\midrule
\multirow{6}{*}{s1.1-7B}
 & \multirow{2}{*}{Uniform}  & Magnitude & 0.1000\textsubscript{\textpm 0.0000} & 0.3468\textsubscript{\textpm 0.0063} & 0.3750\textsubscript{\textpm 0.0354} & 0.6360\textsubscript{\textpm 0.0251} \\
 &                           & Wanda     & 0.1775\textsubscript{\textpm 0.0159} & \textbf{0.3933\textsubscript{\textpm 0.0111}} & 0.5500\textsubscript{\textpm 0.0204} & \textbf{0.8107\textsubscript{\textpm 0.0094}} \\
\cmidrule{2-7}
 & \multirow{2}{*}{OWL}      & Magnitude & 0.0778\textsubscript{\textpm 0.0157} & 0.3468\textsubscript{\textpm 0.0133} & 0.4750\textsubscript{\textpm 0.0204} & 0.6860\textsubscript{\textpm 0.0128} \\
 &                           & Wanda     & \textbf{0.1889\textsubscript{\textpm 0.0157}} & 0.3956\textsubscript{\textpm 0.0119} & 0.5250\textsubscript{\textpm 0.0354} & 0.7967\textsubscript{\textpm 0.0009} \\
\cmidrule{2-7}
 & \multirow{2}{*}{LayerIF}  & Magnitude & 0.1556\textsubscript{\textpm 0.0416} & 0.3687\textsubscript{\textpm 0.0311} & 0.5250\textsubscript{\textpm 0.0204} & 0.8007\textsubscript{\textpm 0.0131} \\
 &                           & Wanda     & 0.1333\textsubscript{\textpm 0.0000} & 0.3956\textsubscript{\textpm 0.0086} & \textbf{0.5583\textsubscript{\textpm 0.0312}} & 0.8013\textsubscript{\textpm 0.0050} \\
\midrule \midrule
\multirow{6}{*}{Qwen3-8B}
 & \multirow{2}{*}{Uniform}  & Magnitude & 0.5889\textsubscript{\textpm 0.0685} & 0.5690\textsubscript{\textpm 0.0212} & \textbf{0.9167\textsubscript{\textpm 0.0236}} & 0.9467\textsubscript{\textpm 0.0081} \\
 &                           & Wanda     & 0.5667\textsubscript{\textpm 0.0272} & \textbf{0.5926\textsubscript{\textpm 0.0195}} & \textbf{0.9167\textsubscript{\textpm 0.0118}} & 0.9533\textsubscript{\textpm 0.0025} \\
\cmidrule{2-7}
 & \multirow{2}{*}{OWL}      & Magnitude & 0.6000\textsubscript{\textpm 0.0816} & 0.5556\textsubscript{\textpm 0.0297} & 0.8917\textsubscript{\textpm 0.0118} & 0.9460\textsubscript{\textpm 0.0016} \\
 &                           & Wanda     & 0.6000\textsubscript{\textpm 0.0720} & 0.5673\textsubscript{\textpm 0.0203} & 0.9000\textsubscript{\textpm 0.0204} & \textbf{0.9573\textsubscript{\textpm 0.0062}} \\
\cmidrule{2-7}
 & \multirow{2}{*}{LayerIF}  & Magnitude & \textbf{0.6222\textsubscript{\textpm 0.0567}} & 0.5774\textsubscript{\textpm 0.0063} & 0.8917\textsubscript{\textpm 0.0312} & 0.9547\textsubscript{\textpm 0.0057} \\
 &                           & Wanda     & \textbf{{0.6222}\textsubscript{\textpm 0.0416}} & 0.5774\textsubscript{\textpm 0.0208} & 0.9083\textsubscript{\textpm 0.0236} & 0.9480\textsubscript{\textpm 0.0043} \\
\bottomrule
\end{tabular}
\label{tab:8192}\vspace{-1mm}
\end{table*}

\subsection{(RQ2) Effect of Layer Sparsity Allocation Strategies on TTS Performance}

%\om {\textbf{Currently writing eveything as if we only have 20\% sparsity, when we get the results for 10\% I will add a small paragraph.}}
\looseness-1\noindent As opposed to uniformly allocating sparsity across layers for the two unstructured pruning strategies, we now explore how state-of-the-art layer-wise sparsity allocation strategies affect the TTS performance of our chosen LLMs. We compare Uniform (i.e. uniform allocation), OWL \cite{pmlr-v235-yin24e}, and LayerIF \cite{askari2025layerif} using a global sparsity of 10\% and 20\% for both unstructured pruning strategies: Magnitude and Wanda. \cref{fig:RQ2} demonstrates TTS performance accuracy of each sparsity allocation strategy, averaged across all four datasets (AIME 24, GPQA-Diamond, AMC 23, and MATH500). Overall, the trends are similar to those observed in RQ1, and hold across all sparsity allocation strategies. It can be observed that several models pruned using unstructured pruning generally end up outperforming their unpruned counterparts.

For s1.1-7B, applying Magnitude-20\% pruning uniformly results in a significant decline in performance relative to the unpruned LLM. In contrast, OWL and LayerIF significantly mitigate this degradation in performance, with LayerIF in particular attaining performance similar to that of the unpruned model. When the pruning ratio is 10\%, we observe that pruning yields better or similar results compared to the unpruned model. For Wanda on s1.1-7B, we can observe that all three sparsity allocation baselines achieve similar performance to the unpruned model with Uniform allocation surpassing all other allocation strategies including the unpruned model. 

Similarly, in the case of Qwen3-8B, all three sparsity allocation strategies outperform the base unpruned model. Interestingly, uniform allocation shows competitive performance when compared with OWL and LayerIF, often matching or slightly exceeding the unpruned LLM across several token budgets. However, OWL allocation achieves better average accuracy across both Wanda and Magnitude pruning methods, and across various thinking token budgets. LayerIF, in general, is not as competitive as OWL for Qwen3-8B but achieves better performance at higher thinking token counts. 

\looseness-1 In \cref{tab:8192}, we provide a more fine-grained analysis for all the sparsity allocation strategies across both LLMs specifically for the maximum token budget case (=8192 thinking tokens). Owing to space constraints, results for other thinking token configurations: 512, 1024, 2048, 4096 are provided in \cref{appendix:Varying Tokens} and show similar trends. Note that as 8192 tokens is the maximum thinking budget, it provides a clear perspective on the models' performance upper-bound. It can be observed that TTS performance for Qwen3-8B across all pruning and allocation strategies shows similar results, indicating that unstructured pruning, regardless of sparsity allocation strategy, is quite effective. For the s1.1-7B model, there is more variability across allocation strategies and pruning approaches. Interestingly, magnitude pruning combined with uniform allocation in this weaker model is more susceptible to performance degradation, but improved allocation strategies such as OWL and LayerIF can mitigate some of these shortcomings. For instance, in Magnitude pruning for AIME24, AMC23, and MATH500, Uniform allocation has accuracies of 0.100, 0.375, 0.636, but  LayerIF attains accuracies of 0.155, 0.525, 0.800, highlighting a substantial improvement. \vspace{1mm}

\looseness-1\noindent \textbf{Remarks on RQ2 Findings.}
Our findings imply that non-uniform sparsity allocation methods, such as OWL and LayerIF, consistently outperform uniform allocation, particularly for the weaker s1 model. While Qwen3-8B is more resilient during TTS even when pruned with uniformly allocated sparsity at higher, more sophisticated allocation strategies provide modest improvements.
Furthermore, while layer-wise sparsity allocation strategies constitute an important parametric choice for unstructured pruning methods, we also conduct additional experiments to observe if the specific choice of pruning layer (MLP or attention) leads to drastically differing TTS performance. Due to space constraints, we provide these additional results in \cref{appendix:isolation} but find that the choice of layer for pruning does not affect TTS performance significantly, unlike the choice of sparsity allocation strategy.

\section{Discussion}

We now discuss some salient points of interest, extrapolating beyond the results we have obtained from our experiments.\vspace{1.5mm}

\looseness-1\noindent \textbf{Unstructured pruning outperforms unpruned LLMs in TTS. }
Our results demonstrating that unstructured pruning not only preserves performance but can even exceed that of the unpruned LLM constitutes an interesting finding. There are several reasons outlined in prior work that help explain this outcome. For instance, the \textit{Lottery Ticket Hypothesis} \cite{frankle2018the} suggests that dense neural networks contain high-performing subnetworks that can outperform the original unpruned model when trained in isolation. In a similar vein, \citet{JMLR:v26:23-0832} theoretically demonstrate that moderate pruning rates can indeed improve generalization bounds. Moreover, \citet{NEURIPS2020_ef2ee09e} find that pruning can improve generalization in deep neural networks especially when post-training is undertaken with the pruned models. Other works have also shown that certain pruned models can maintain performance comparable to their unpruned counterparts even at high sparsity levels (of around 50\%), without requiring post-training \cite{sun2024a,pmlr-v202-frantar23a}. \vspace{1.5mm}

\looseness-1\noindent \textbf{Qualitative analysis of reasoning chains.} Complementary to these works, our findings indicate that, even for TTS and reasoning, pruning can maintain or exceed the performance of the unpruned model, without requiring post-pruning training. Additionally, our findings show that unstructured pruning, which masks specific weights rather than entire blocks, can achieve this effect while preserving the model's overall reasoning capability. Thus, not all parameters seem to be equally important for reasoning, implying that carefully applied pruning can serve as a method to enhance downstream TTS performance. To analyze this further, we first provide qualitative examples in Appendix \ref{appendix:Qualitative} that demonstrate improved reasoning chains after pruning (for Wanda-Uniform 20\%) compared to the base model. Second, we utilize GPT-4o as a Judge LLM and find that it prefers the reasoning traces for this pruned model more than the unpruned base LLM. These results are provided in Appendix \ref{appendix:Quantitative} due to space limitations.\vspace{1.5mm}

\looseness-1\noindent \textbf{Pruning layers selected by structured pruning differ significantly compared to unstructured layer-wise sparsity allocation strategies.} To interpret our findings further, we seek to analyze which layers are selected for pruning across structured (ShortGPT) and unstructured (specifically, layer-sparsity allocation strategies such as OWL and LayerIF, since the default unstructured pruning strategy is just to uniformly allocate rates) methods. For s1.1-7B, ShortGPT completely removes Layers 16 and 17, while OWL prunes Layer 10 and 11 the most, and LayerIF applies the highest pruning to Layers 11 and 12. Similarly, for Qwen3-8B, the layers removed via ShortGPT (20 and 17) differ from the layers selected by both OWL (8 and 11) and LayerIF (0 and 4). This distinction in the layers being selected gives an indication as to why performance differs across the methods, as there is little to no overlap across layers being pruned across these methods. %Another key distinction between structured and unstructured pruning is that, in the latter, even the most pruned layers do not deviate substantially from the target global sparsity($+5\%$ at most).
\vspace{1.5mm}

\noindent \textbf{Non-uniform sparsity allocation can augment TTS performance when pruning has adverse effects.}
We find that for less performant models such as s1.1-7B, certain pruning strategies are more detrimental than others, i.e., Magnitude-Uniform pruning at 20\% sparsity lags behind Wanda-Uniform at 20\%. However, interestingly, non-uniform sparsity methods, such as LayerIF, can significantly reduce this degradation. This indicates that while Magnitude pruning alone may remove weights without considering their functional importance, allocating sparsity in a layer-aware or influence-aware manner helps preserve critical components of the model’s reasoning capacity.\vspace{1.5mm}

\noindent \textbf{Uniform sparsity allocation is a strong default option for sparsity allocation.}
As we have observed in the previous section, for performant TTS-enabled reasoning LLMs such as Qwen3-8B, when using more advanced pruning methods such as Wanda, even the simplest sparsity allocation strategy, i.e. uniform allocation, remains highly competitive compared to OWL and LayerIF across most thinking token budgets. This suggests that higher-capacity models are inherently more robust to sparsity allocation patterns, likely due to their greater parameter redundancy. Hence, carefully allocating sparsity across layers to improve performance leads to less pronounced effects for high-performing LLMs and indicates that simple uniform pruning may already be sufficient for maintaining or, on occasions exceeding reasoning performance.

\section{Conclusion}
We systematically explored the impact of pruning on reasoning performance under test-time scaling across multiple benchmarks and reasoning LLMs. In agreement with prior work, we found that structured pruning is detrimental for TTS performance across thinking token budgets. However, our findings contrast for the relatively unexplored unstructured pruning setting as our results demonstrate that unstructured pruning can consistently surpass the unpruned LLMs across various datasets and pruning strategies. We also conducted analysis across different layer-aware sparsity allocation strategies for unstructured pruning and find that non-uniform strategies consistently outperform uniform allocation for simpler models but the difference between uniform and non-uniform strategies is minimal for more performant LLMs. Our work seeks to challenge the conventional notion that pruning strategies are inherently detrimental for TTS and galvanize future research into the role pruning can play in augmenting LLM TTS performance.

%\clearpage

\section*{Limitations}
While we believe our experiments consider a diverse range of pruning scenarios, there are still limitations and unexplored areas that future work might address. To ensure fairness in comparison with other non-training baselines we considered, we did not explore pruning methods, e.g. SparseGPT \cite{pmlr-v202-frantar23a}, that require training on data post the pruning process, although these can be investigated in future work. Moreover, while outside the scope of our work since our study was localized to pruning, the effect of other compression techniques such as \textit{quantization} \cite{egashira2024exploiting, liu2024spinquant} and how they impact downstream TTS performance would constitute an interesting research direction as well. Finally, recent work has outlined safety issues with TTS \cite{nahin2025less} and a complementary research direction could investigate if pruning can serve a role in mitigating these given our findings that pruning can even surpass the limits of the unpruned model.

%Our paper focuses on sequential test-time allocation strategies; parallel approaches can be explored and evaluated. Additionally, our analysis is limited to mid-sized open-source models (s1.1-7B and Qwen3-8B). While larger models with massive parameter counts (e.g., 70B) may respond differently to pruning, evaluating them is outside the scope of this study due to resource constraints

\section*{Ethics Statement}
\looseness-1 Our work explores the effectiveness of pruning methods for LLMs during TTS. All the datasets and models used in this study are publicly available and widely used. We do not introduce any new data, sensitive, or personal information, and focus solely on model performance in TTS. Our findings aim to enhance LLM robustness and performance in TTS and do not generate or promote harmful content.

%\clearpage
\bibliography{refs}
%\clearpage

\appendix
%\onecolumn
\section*{Appendix}

\section{Implementation Details}
\label{appendix:Implementation}
All experimental settings reported in our paper, including pruning methods and sparsity allocation strategies, follow the standard methodologies and implementation details described in the original papers. Only minor adjustments were introduced where necessary to ensure compatibility with Qwen-based architectures and their corresponding model configurations. The datasets used in this work are obtained directly from their standard and publicly available HuggingFace repositories, without additional modifications unless explicitly stated. For generating responses from the models during evaluation, we maintain a fixed sampling temperature of 1 across all experiments to ensure consistency in generation behavior. 

To account for the randomness in the generation process and to provide more consistent outputs, we evaluate each experiment using three different random seeds, specifically 7, 11, and 42. All model evaluations and benchmark measurements are conducted using the lm-evaluation-harness framework, which provides a standardized evaluation pipeline for large language models and ensures reproducibility and consistency across different experimental settings.

{For Wanda, we use the default settings given in the Wanda repository, and as is standard, our calibration data is a subset of the C4 data. Specifically, our calibration dataset consists of 128 samples drawn from the English subset of the C4 dataset. For LayerIF \cite{askari2025layerif}, we adopt the TracIN \cite{NEURIPS2020_e6385d39} formulation to estimate layer-wise influence by computing two backward passes on the S1 dataset, one over the chain-of-thought reasoning and one over the answer, using cross-entropy loss, with the resulting per-layer gradient dot products guiding the allocation of sparsity across layers.}

\section{Unstructured Pruning Layer Choice (Attention vs MLP) Effects on TTS Performance}
\label{appendix:isolation}
\begin{figure*}[t]
    \centering
    \includegraphics[width=\textwidth]{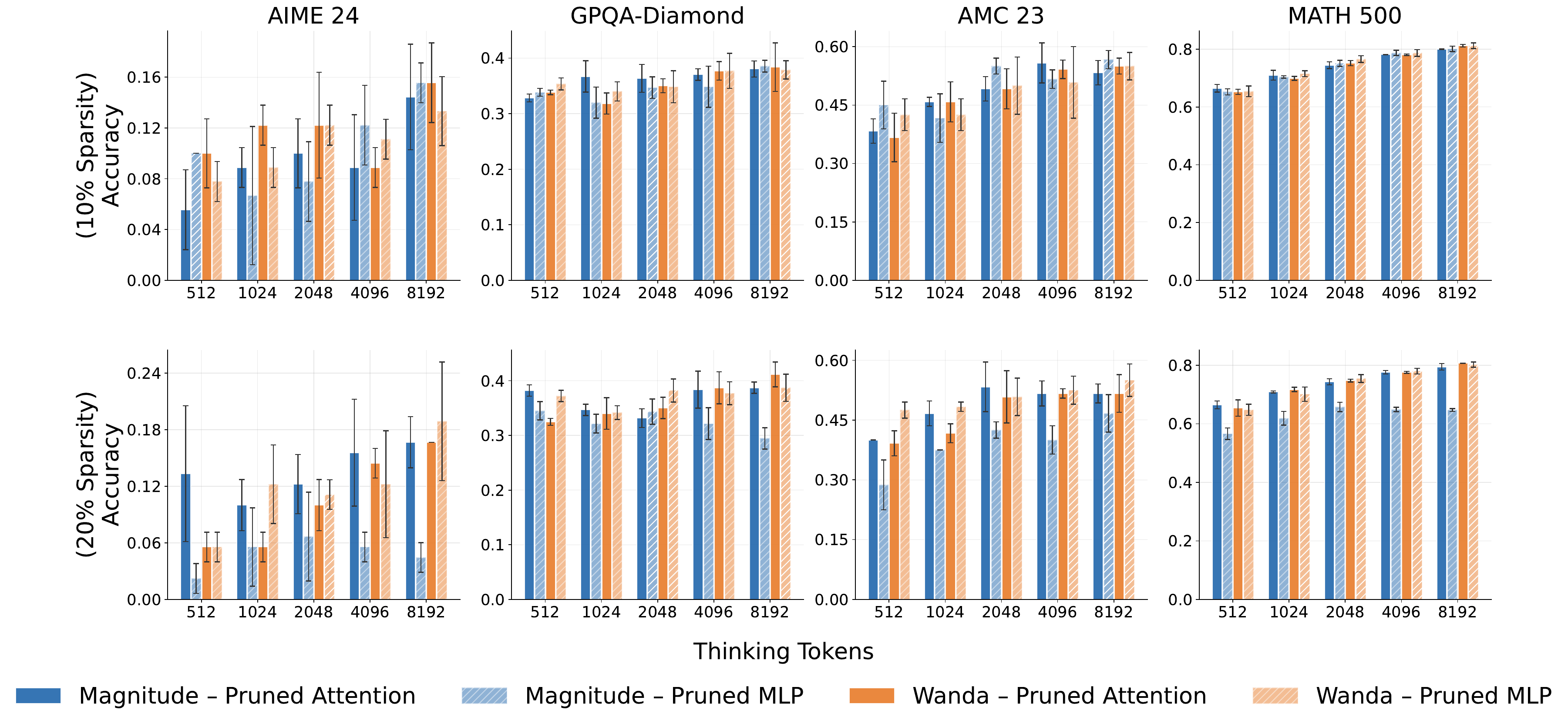}
    \caption{s1.1-7B results, when attention and feed-forward MLP layers are pruned in isolation at different sparsity ratios (10\% and 20\%)  using Magnitude and Wanda with uniform sparsity across all four of our selected datasets and all five thinking token budgets.}
    \label{fig:APP2}
\end{figure*}

\begin{figure*}[t]
    \centering
    \includegraphics[width=\textwidth]{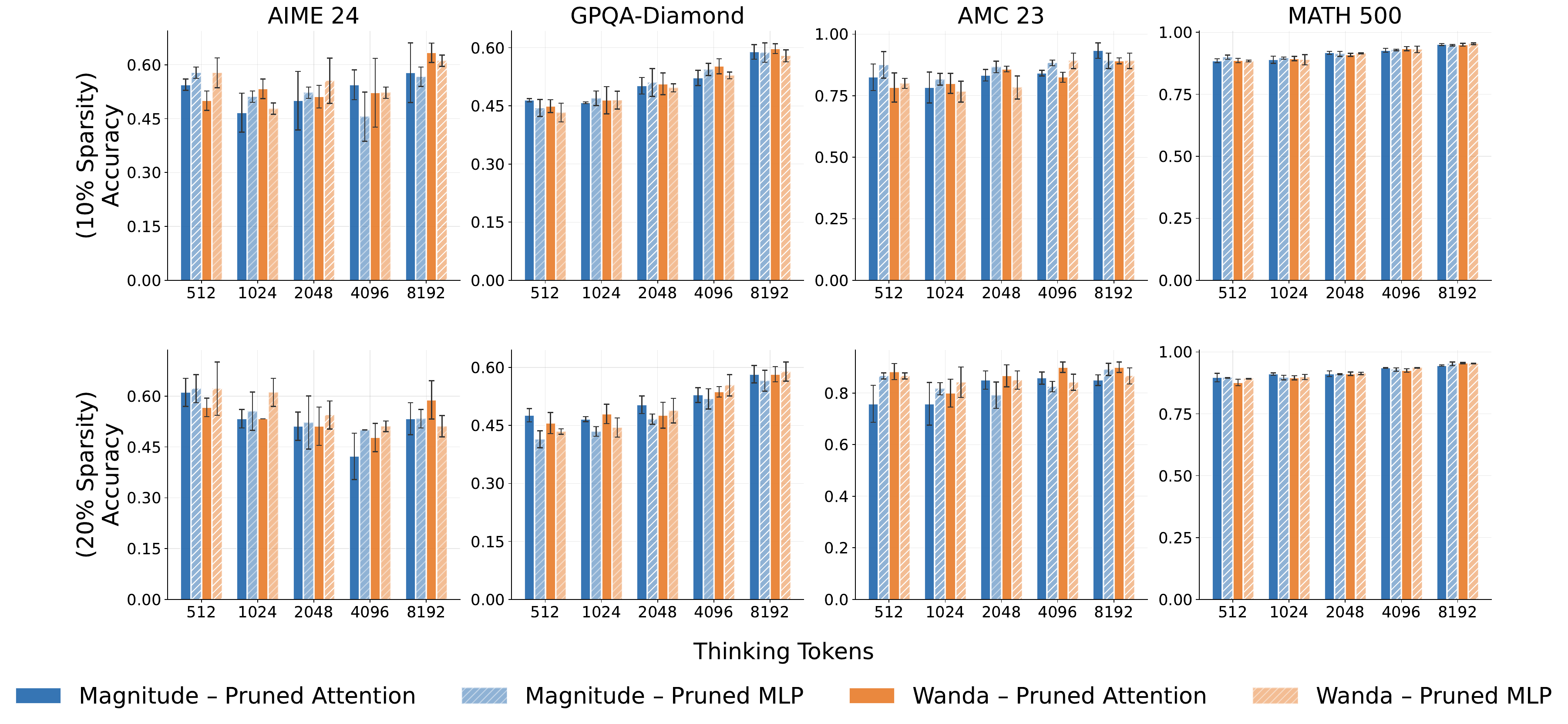}
    \caption{Qwen3-8B results, when attention and feed-forward MLP layers are pruned in isolation at different sparsity ratios (10\% and 20\%)  using Magnitude and Wanda with uniform sparsity across all four of our selected datasets and all five thinking token budgets.}
    \label{fig:APP1}
\end{figure*}

Examining both \cref{fig:APP2} and \cref{fig:APP1}, we do not observe any clear overarching pattern or consistent trend across the experiments. Nevertheless, these experiments yield specific results that can be meaningfully extrapolated. For instance, in \cref{fig:APP2}, Magnitude-Pruned MLP typically outperforms Magnitude-Pruned Attention at 10\% sparsity, but this relationship is inverted when sparsity increases to 20\%. This inversion likely reflects the higher redundancy tolerance of MLP layers at low sparsity, which might diminish if pruned over a critical threshold. 

Overall, there remains some variability across pruned layer choices for s1.1-7B. However, as shown in \cref{fig:APP1}, for the more performant Qwen3-8B model this variability is much smaller, suggesting that the choice of which layer type to prune does not lead to substantial differences overall especially if performance is already very good. Taken together, these observations indicate that while layer selection can introduce very minor fluctuations, the overall impact remains relatively limited across the evaluated settings.

\section{Additional Results for Varying Thinking Tokens}
% for rq2 - similar tables as the table in the main paper
% ── 512 tokens ──
\label{appendix:Varying Tokens}
We present the fine-grained results for all five thinking tokens configurations (512, 1024, 2048, 4096, and 8192) under the  10\% global sparsity setting in \cref{tab:10_512,tab:10_1024,tab:10_2048,tab:10_4096,tab:10_8192}.  In addition, we report the corresponding results for the remaining four thinking token configurations (512, 1024, 2048, and 4096) under the 20\% sparsity setting in \cref{tab:512_20,tab:1024_20,tab:2048_20,tab:4096_20}. This detailed breakdown allows us to more closely examine how different thinking token budgets interact with pruning and sparsity levels, providing a clearer view of the performance trends across configurations.

\section{Qualitative Analysis of Chain-of-Thoughts for Pruned vs Unpruned Models}
% for rq2 - similar tables as the table in the main paper
% ── 512 tokens ──
\label{appendix:Qualitative}
The qualitative examples for the unpruned and Wanda-Uniform 20\% pruned variant of the Qwen3-8B models are provided in \cref{tab:cot_comparison,tab:cot_comparison_amc,tab:cot_comparison_gpqa,tab:cot_comparison_math}. Where two qualitative examples for each of the four datasets (AIME24, AMC23, GPQA-Diamond, and MATH500) at 2048 thinking tokens are shown. Overall, inspection of the chain of thought reveals no clearly identifiable signal. The chain of thought generation in both cases shows coherent outputs. Interestingly, \cite{wang2025fewer} shows that the chain-of-thought generated by \textit{structurally} pruned models often contains incoherent outputs and repeated segments; issues that are not observed in either unpruned models or those pruned using \textit{unstructured} methods. This finding is consistent with the quantitative results, where unstructured pruned models often match or surpass the performance of their unpruned counterparts.

\section{LLM as a judge Analysis of Chain-of-Thoughts for Pruned vs Unpruned Models}
\label{appendix:Quantitative}

{To complement the qualitative chain-of-thought analysis in Appendix \ref{appendix:Qualitative}, we conducted an auxiliary evaluation of the Chain-of-Thought outputs for AIME24 using an LLM as a Judge (GPT-4o), comparing unpruned and Wanda-Uniform 20\% for the same Qwen3-8B model with 2048 and 8192 token budgets, with results in Table \ref{tab:llm_judge}. The results indicate that, although the differences are modest, the Wanda 20\% pruned LLM consistently achieves higher judge scores than the unpruned base model. Both the qualitative and LLM-as-a-Judge evaluations point to the same conclusion: unstructured pruning preserves the integrity of the chain-of-thought generation even at 20\% sparsity.}

\section{Comparison of Structured and Unstructured Pruning at the Same Sparsity Rates}
\label{appendix:Same_Sparsity}
{Throughout this paper, we followed common practice in unstructured pruning literature and reported results at 10\% and 20\% sparsity, which are standard sparsity rates for pruning. The goal was to show that even at these levels, unstructured pruning significantly outperforms structured pruning settings that remove at most 7\% of parameters. To provide a comparison under a more matched parameter budget, we include additional results under pruning budgets for both structured and unstructured pruning on AIME24 and GPQA-Diamond in Table \ref{tab:qwen3_results_equal} and \ref{tab:s1_results_equal}. Specifically, we report results for Qwen3-8B at 2.7\% and 5.7\% sparsity (equivalent to removing 1 and 2 layers, respectively) and s1.1-7B at 3.5\% and 7.0\% sparsity (also corresponding to removing 1 and 2 layers).

The results presented in Table \ref{tab:qwen3_results_equal} and \ref{tab:s1_results_equal} reinforce our findings from RQ1. Across both models and datasets, the best performing variant at each token budget is either the unpruned model or one of the unstructured pruning methods, while structured pruning via ShortGPT exhibits substantial performance degradation in comparison. This holds regardless of whether one or two layers are removed, suggesting that even modest structural pruning is disproportionately harmful to reasoning performance relative to the equivalent amount of unstructured sparsity.}

\section{Additional Results on Llama-3.1-8B}\label{appendix:Additional Results}

Additionally, we evaluate DeepSeek-R1-Distill-Llama-8B\footnote{\url{https://huggingface.co/deepseek-ai/DeepSeek-R1-Distill-Llama-8B}}, a Llama-3.1-8B LLM trained using reasoning traces generated by DeepSeek-R1 to enable reasoning and TTS. As can be seen in Table \ref{tab:llama_results}, our TTS-pruning findings generalize to this LLM as well. The performance is assessed under the same experimental setup as RQ1 on AIME24 and GPQA-Diamond.

\section{Code and Reproducibility}

Experiments for this paper were conducted on two computing clusters: one equipped with RTX A6000 GPUs and the other with B200 GPUs.
The code will be released shortly via a public GitHub repository. %We provide the code and instructions to reproduce our results in this repository: \url{https://anonymous.4open.science/r/TTS-Pruning-6C13}.

 %, except for the implementation of LayerIF.

% For LayerIF, we make targeted changes in how we calculate layer-wise influence and what layers to allocate for the TTS setting. The original LayerIF paper only considers sparsity allocation across attention layers. This setting does not provide a meaningful comparison with other methods, since attention layers constitute only a small fraction of the total model parameters, making it infeasible to achieve 10\% or 20\% global sparsity through attention pruning alone. Thus, we calculate layer-wise influence across not only attention layers but also feed-forward MLP layers. Therefore, we compute layer-wise influence across both attention layers and feed-forward MLP layers.  Unlike the original LayerIF, since we consider both attention and MLP layers, we do not compute the Hessian for influence calculation; instead, we compute influence via TracIN, as computing the full Hessian for the entire model would be computationally impractical.

% 2 figures - each model

\begin{table*}[t]
\centering
\small
\caption{Results for s1.1-7B and Qwen3-8B across allocation strategies and pruning approaches (Global Sparsity: 20\%, Max Thinking Tokens: 512).}
\begin{tabular}{lll|cccc}
\toprule
\textbf{Model} & \textbf{Allocation} & \textbf{Pruning} & \textbf{AIME24} & \textbf{GPQA-Diamond} & \textbf{AMC23} & \textbf{MATH500} \\
\midrule
\multirow{6}{*}{s1.1-7B}
 & \multirow{2}{*}{Uniform}  & Magnitude & 0.0222\textsubscript{±0.0314} & 0.2963\textsubscript{±0.0086} & 0.3083\textsubscript{±0.0471} & 0.5673\textsubscript{±0.0198} \\
 &                           & Wanda     & 0.0551\textsubscript{±0.0154} & 0.3463\textsubscript{±0.0118} & 0.4333\textsubscript{±0.0312} & 0.6420\textsubscript{±0.0247} \\
\cmidrule{2-7}
 & \multirow{2}{*}{OWL}      & Magnitude & 0.0333\textsubscript{±0.0000} & 0.3114\textsubscript{±0.0252} & 0.3667\textsubscript{±0.0943} & 0.5720\textsubscript{±0.0295} \\
 &                           & Wanda     & 0.1111\textsubscript{±0.0314} & 0.3620\textsubscript{±0.0227} & 0.4167\textsubscript{±0.0514} & 0.6447\textsubscript{±0.0219} \\
\cmidrule{2-7}
 & \multirow{2}{*}{LayerIF}  & Magnitude & 0.1000\textsubscript{±0.0471} & 0.3569\textsubscript{±0.0104} & 0.3917\textsubscript{±0.0425} & 0.6380\textsubscript{±0.0113} \\
 &                           & Wanda     & 0.1000\textsubscript{±0.0000} & 0.3418\textsubscript{±0.0133} & 0.4500\textsubscript{±0.0354} & 0.6460\textsubscript{±0.0227} \\
\midrule \midrule
\multirow{6}{*}{Qwen3-8B}
 & \multirow{2}{*}{Uniform}  & Magnitude & 0.6000\textsubscript{±0.0943} & 0.4512\textsubscript{±0.0265} & 0.8750\textsubscript{±0.0354} & 0.9207\textsubscript{±0.0120} \\
 &                           & Wanda     & 0.6222\textsubscript{±0.0416} & 0.4680\textsubscript{±0.0304} & 0.8833\textsubscript{±0.0236} & 0.8920\textsubscript{±0.0118} \\
\cmidrule{2-7}
 & \multirow{2}{*}{OWL}      & Magnitude & 0.6556\textsubscript{±0.0314} & 0.4495\textsubscript{±0.0071} & 0.8917\textsubscript{±0.0425} & 0.9267\textsubscript{±0.0019} \\
 &                           & Wanda     & 0.6111\textsubscript{±0.0157} & 0.4512\textsubscript{±0.0086} & 0.9000\textsubscript{±0.0204} & 0.9040\textsubscript{±0.0057} \\
\cmidrule{2-7}
 & \multirow{2}{*}{LayerIF}  & Magnitude & 0.6000\textsubscript{±0.0471} & 0.4276\textsubscript{±0.0126} & 0.8167\textsubscript{±0.0656} & 0.9160\textsubscript{±0.0114} \\
 &                           & Wanda     & 0.5778\textsubscript{±0.0416} & 0.4411\textsubscript{±0.0203} & 0.8333\textsubscript{±0.0425} & 0.8827\textsubscript{±0.0019} \\
\bottomrule
\end{tabular}
\label{tab:512_20}
\end{table*}

% ── 1024 tokens ──
\begin{table*}[t]
\centering
\small
\caption{Results for s1.1-7B and Qwen3-8B across allocation strategies and pruning approaches (Global Sparsity: 20\%, Max Thinking Tokens: 1024).}
\begin{tabular}{lll|cccc}
\toprule
\textbf{Model} & \textbf{Allocation} & \textbf{Pruning} & \textbf{AIME24} & \textbf{GPQA-Diamond} & \textbf{AMC23} & \textbf{MATH500} \\
\midrule
\multirow{6}{*}{s1.1-7B}
 & \multirow{2}{*}{Uniform}  & Magnitude & 0.0110\textsubscript{±0.0156} & 0.3266\textsubscript{±0.0227} & 0.3667\textsubscript{±0.0312} & 0.6133\textsubscript{±0.0196} \\
 &                           & Wanda     & 0.0778\textsubscript{±0.0314} & 0.3463\textsubscript{±0.0118} & 0.4583\textsubscript{±0.0773} & 0.7080\textsubscript{±0.0193} \\
\cmidrule{2-7}
 & \multirow{2}{*}{OWL}      & Magnitude & 0.0556\textsubscript{±0.0157} & 0.3384\textsubscript{±0.0251} & 0.4333\textsubscript{±0.0471} & 0.6227\textsubscript{±0.0106} \\
 &                           & Wanda     & 0.0778\textsubscript{±0.0157} & 0.3283\textsubscript{±0.0041} & 0.5250\textsubscript{±0.0000} & 0.7087\textsubscript{±0.0139} \\
\cmidrule{2-7}
 & \multirow{2}{*}{LayerIF}  & Magnitude & 0.0667\textsubscript{±0.0000} & 0.3620\textsubscript{±0.0104} & 0.4583\textsubscript{±0.0118} & 0.6860\textsubscript{±0.0140} \\
 &                           & Wanda     & 0.0778\textsubscript{±0.0157} & 0.3468\textsubscript{±0.0304} & 0.4750\textsubscript{±0.0540} & 0.7107\textsubscript{±0.0157} \\
\midrule \midrule
\multirow{6}{*}{Qwen3-8B}
 & \multirow{2}{*}{Uniform}  & Magnitude & 0.6667\textsubscript{±0.0272} & 0.4293\textsubscript{±0.0189} & 0.8500\textsubscript{±0.0354} & 0.9247\textsubscript{±0.0124} \\
 &                           & Wanda     & 0.5000\textsubscript{±0.0720} & 0.4613\textsubscript{±0.0104} & 0.8250\textsubscript{±0.0540} & 0.9220\textsubscript{±0.0140} \\
\cmidrule{2-7}
 & \multirow{2}{*}{OWL}      & Magnitude & 0.6000\textsubscript{±0.0544} & 0.4630\textsubscript{±0.0063} & 0.8917\textsubscript{±0.0471} & 0.9247\textsubscript{±0.0041} \\
 &                           & Wanda     & 0.4889\textsubscript{±0.0786} & 0.4478\textsubscript{±0.0063} & 0.8750\textsubscript{±0.0540} & 0.9080\textsubscript{±0.0059} \\
\cmidrule{2-7}
 & \multirow{2}{*}{LayerIF}  & Magnitude & 0.6333\textsubscript{±0.0471} & 0.4091\textsubscript{±0.0071} & 0.8417\textsubscript{±0.0236} & 0.9227\textsubscript{±0.0074} \\
 &                           & Wanda     & 0.5000\textsubscript{±0.0720} & 0.4663\textsubscript{±0.0063} & 0.8250\textsubscript{±0.0000} & 0.9047\textsubscript{±0.0164} \\
\bottomrule
\end{tabular}
\label{tab:1024_20}
\end{table*}

% ── 2048 tokens ──
\begin{table*}[t]
\centering
\small
\caption{Results for s1.1-7B and Qwen3-8B across allocation strategies and pruning approaches (Global Sparsity: 20\%, Max Thinking Tokens: 2048).}
\begin{tabular}{lll|cccc}
\toprule
\textbf{Model} & \textbf{Allocation} & \textbf{Pruning} & \textbf{AIME24} & \textbf{GPQA-Diamond} & \textbf{AMC23} & \textbf{MATH500} \\
\midrule
\multirow{6}{*}{s1.1-7B}
 & \multirow{2}{*}{Uniform}  & Magnitude & 0.0332\textsubscript{±0.0272} & 0.3418\textsubscript{±0.0024} & 0.4333\textsubscript{±0.0589} & 0.6447\textsubscript{±0.0082} \\
 &                           & Wanda     & 0.1332\textsubscript{±0.0001} & 0.3697\textsubscript{±0.0295} & 0.5333\textsubscript{±0.0312} & 0.7493\textsubscript{±0.0077} \\
\cmidrule{2-7}
 & \multirow{2}{*}{OWL}      & Magnitude & 0.0778\textsubscript{±0.0157} & 0.3333\textsubscript{±0.0082} & 0.4083\textsubscript{±0.0656} & 0.6720\textsubscript{±0.0059} \\
 &                           & Wanda     & 0.1111\textsubscript{±0.0416} & 0.3603\textsubscript{±0.0195} & 0.5000\textsubscript{±0.0354} & 0.7413\textsubscript{±0.0184} \\
\cmidrule{2-7}
 & \multirow{2}{*}{LayerIF}  & Magnitude & 0.0889\textsubscript{±0.0314} & 0.3822\textsubscript{±0.0063} & 0.4667\textsubscript{±0.0118} & 0.7433\textsubscript{±0.0217} \\
 &                           & Wanda     & 0.0889\textsubscript{±0.0157} & 0.3670\textsubscript{±0.0104} & 0.5083\textsubscript{±0.0236} & 0.7540\textsubscript{±0.0086} \\
\midrule \midrule
\multirow{6}{*}{Qwen3-8B}
 & \multirow{2}{*}{Uniform}  & Magnitude & 0.5556\textsubscript{±0.0685} & 0.4949\textsubscript{±0.0041} & 0.8833\textsubscript{±0.0118} & 0.9320\textsubscript{±0.0028} \\
 &                           & Wanda     & 0.5778\textsubscript{±0.0314} & 0.4933\textsubscript{±0.0024} & 0.8667\textsubscript{±0.0514} & 0.9180\textsubscript{±0.0033} \\
\cmidrule{2-7}
 & \multirow{2}{*}{OWL}      & Magnitude & 0.5444\textsubscript{±0.0157} & 0.4949\textsubscript{±0.0109} & 0.9083\textsubscript{±0.0514} & 0.9340\textsubscript{±0.0016} \\
 &                           & Wanda     & 0.5333\textsubscript{±0.0544} & 0.5017\textsubscript{±0.0298} & 0.8833\textsubscript{±0.0312} & 0.9213\textsubscript{±0.0019} \\
\cmidrule{2-7}
 & \multirow{2}{*}{LayerIF}  & Magnitude & 0.5556\textsubscript{±0.0567} & 0.4949\textsubscript{±0.0149} & 0.8250\textsubscript{±0.0354} & 0.9253\textsubscript{±0.0041} \\
 &                           & Wanda     & 0.5444\textsubscript{±0.0416} & 0.4865\textsubscript{±0.0156} & 0.8083\textsubscript{±0.0118} & 0.9147\textsubscript{±0.0041} \\
\bottomrule
\end{tabular}
\label{tab:2048_20}
\end{table*}

% ── 4096 tokens ──
\begin{table*}[t]
\centering
\small
\caption{Results for s1.1-7B and Qwen3-8B across allocation strategies and pruning approaches (Global Sparsity: 20\%, Max Thinking Tokens: 4096).}
\begin{tabular}{lll|cccc}
\toprule
\textbf{Model} & \textbf{Allocation} & \textbf{Pruning} & \textbf{AIME24} & \textbf{GPQA-Diamond} & \textbf{AMC23} & \textbf{MATH500} \\
\midrule
\multirow{6}{*}{s1.1-7B}
 & \multirow{2}{*}{Uniform}  & Magnitude & 0.0110\textsubscript{±0.0156} & 0.3401\textsubscript{±0.0104} & 0.3750\textsubscript{±0.0736} & 0.6380\textsubscript{±0.0145} \\
 &                           & Wanda     & 0.1885\textsubscript{±0.0314} & 0.3883\textsubscript{±0.0181} & 0.5333\textsubscript{±0.0236} & 0.7713\textsubscript{±0.0068} \\
\cmidrule{2-7}
 & \multirow{2}{*}{OWL}      & Magnitude & 0.0778\textsubscript{±0.0157} & 0.3670\textsubscript{±0.0126} & 0.4167\textsubscript{±0.0425} & 0.6927\textsubscript{±0.0094} \\
 &                           & Wanda     & 0.1000\textsubscript{±0.0544} & 0.3956\textsubscript{±0.0437} & 0.5333\textsubscript{±0.0236} & 0.7727\textsubscript{±0.0057} \\
\cmidrule{2-7}
 & \multirow{2}{*}{LayerIF}  & Magnitude & 0.1333\textsubscript{±0.0272} & 0.3872\textsubscript{±0.0423} & 0.4833\textsubscript{±0.0312} & 0.7720\textsubscript{±0.0134} \\
 &                           & Wanda     & 0.1333\textsubscript{±0.0272} & 0.3838\textsubscript{±0.0165} & 0.5083\textsubscript{±0.0425} & 0.7880\textsubscript{±0.0059} \\
\midrule \midrule
\multirow{6}{*}{Qwen3-8B}
 & \multirow{2}{*}{Uniform}  & Magnitude & 0.5000\textsubscript{±0.0471} & 0.5067\textsubscript{±0.0298} & 0.8417\textsubscript{±0.0425} & 0.9320\textsubscript{±0.0028} \\
 &                           & Wanda     & 0.5000\textsubscript{±0.0272} & 0.5455\textsubscript{±0.0082} & 0.8667\textsubscript{±0.0236} & 0.9280\textsubscript{±0.0028} \\
\cmidrule{2-7}
 & \multirow{2}{*}{OWL}      & Magnitude & 0.5556\textsubscript{±0.0314} & 0.5017\textsubscript{±0.0133} & 0.8750\textsubscript{±0.0408} & 0.9280\textsubscript{±0.0059} \\
 &                           & Wanda     & 0.5111\textsubscript{±0.0416} & 0.5236\textsubscript{±0.0227} & 0.8750\textsubscript{±0.0204} & 0.9327\textsubscript{±0.0082} \\
\cmidrule{2-7}
 & \multirow{2}{*}{LayerIF}  & Magnitude & 0.5333\textsubscript{±0.0272} & 0.5101\textsubscript{±0.0041} & 0.8667\textsubscript{±0.0312} & 0.9407\textsubscript{±0.0077} \\
 &                           & Wanda     & 0.5111\textsubscript{±0.0831} & 0.5404\textsubscript{±0.0251} & 0.8583\textsubscript{±0.0312} & 0.9267\textsubscript{±0.0047} \\
\bottomrule
\end{tabular}
\label{tab:4096_20}
\end{table*}

% ── 512 tokens ──
\begin{table*}[t]
\centering
\small
\caption{Results for s1.1-7B and Qwen3-8B across allocation strategies and pruning approaches (Global Sparsity: 10\%, Max Thinking Tokens: 512).}
\begin{tabular}{lll|cccc}
\toprule
\textbf{Model} & \textbf{Allocation} & \textbf{Pruning} & \textbf{AIME24} & \textbf{GPQA-Diamond} & \textbf{AMC23} & \textbf{MATH500} \\
\midrule
\multirow{6}{*}{s1.1-7B}
 & \multirow{2}{*}{Uniform}  & Magnitude & 0.1332\textsubscript{±0.0272} & 0.3246\textsubscript{±0.0233} & 0.3917\textsubscript{±0.0471} & 0.6660\textsubscript{±0.0140} \\
 &                           & Wanda     & 0.0775\textsubscript{±0.0416} & 0.3447\textsubscript{±0.0118} & 0.4000\textsubscript{±0.0408} & 0.6480\textsubscript{±0.0043} \\
\cmidrule{2-7}
 & \multirow{2}{*}{OWL}      & Magnitude & 0.0667\textsubscript{±0.0272} & 0.3283\textsubscript{±0.0143} & 0.3917\textsubscript{±0.0425} & 0.6680\textsubscript{±0.0102} \\
 &                           & Wanda     & 0.0333\textsubscript{±0.0272} & 0.3316\textsubscript{±0.0265} & 0.4417\textsubscript{±0.0312} & 0.6487\textsubscript{±0.0175} \\
\cmidrule{2-7}
 & \multirow{2}{*}{LayerIF}  & Magnitude & 0.1000\textsubscript{±0.0272} & 0.3418\textsubscript{±0.0212} & 0.3750\textsubscript{±0.0408} & 0.6547\textsubscript{±0.0139} \\
 &                           & Wanda     & 0.0667\textsubscript{±0.0272} & 0.3485\textsubscript{±0.0214} & 0.4167\textsubscript{±0.0514} & 0.6533\textsubscript{±0.0077} \\
\midrule \midrule
\multirow{6}{*}{Qwen3-8B}
 & \multirow{2}{*}{Uniform}  & Magnitude & 0.5889\textsubscript{±0.0157} & 0.4529\textsubscript{±0.0086} & 0.8500\textsubscript{±0.0354} & 0.8927\textsubscript{±0.0062} \\
 &                           & Wanda     & 0.6000\textsubscript{±0.0272} & 0.4343\textsubscript{±0.0143} & 0.8500\textsubscript{±0.0204} & 0.8760\textsubscript{±0.0075} \\
\cmidrule{2-7}
 & \multirow{2}{*}{OWL}      & Magnitude & 0.6111\textsubscript{±0.0416} & 0.4461\textsubscript{±0.0104} & 0.8750\textsubscript{±0.0000} & 0.9087\textsubscript{±0.0075} \\
 &                           & Wanda     & 0.5333\textsubscript{±0.0471} & 0.4276\textsubscript{±0.0104} & 0.8333\textsubscript{±0.0312} & 0.8980\textsubscript{±0.0043} \\
\cmidrule{2-7}
 & \multirow{2}{*}{LayerIF}  & Magnitude & 0.6889\textsubscript{±0.0157} & 0.4478\textsubscript{±0.0298} & 0.8250\textsubscript{±0.0736} & 0.9020\textsubscript{±0.0075} \\
 &                           & Wanda     & 0.5889\textsubscript{±0.0685} & 0.4461\textsubscript{±0.0195} & 0.8667\textsubscript{±0.0236} & 0.8873\textsubscript{±0.0062} \\
\bottomrule
\end{tabular}
\label{tab:10_512}
\end{table*}

% ── 1024 tokens ──
\begin{table*}[t]
\centering
\small
\caption{Results for s1.1-7B and Qwen3-8B across allocation strategies and pruning approaches (Global Sparsity: 10\%, Max Thinking Tokens: 1024).}
\begin{tabular}{lll|cccc}
\toprule
\textbf{Model} & \textbf{Allocation} & \textbf{Pruning} & \textbf{AIME24} & \textbf{GPQA-Diamond} & \textbf{AMC23} & \textbf{MATH500} \\
\midrule
\multirow{6}{*}{s1.1-7B}
 & \multirow{2}{*}{Uniform}  & Magnitude & 0.0554\textsubscript{±0.0315} & 0.3398\textsubscript{±0.0048} & 0.4333\textsubscript{±0.0118} & 0.7093\textsubscript{±0.0137} \\
 &                           & Wanda     & 0.0773\textsubscript{±0.0160} & 0.3293\textsubscript{±0.0300} & 0.4083\textsubscript{±0.0425} & 0.6953\textsubscript{±0.0165} \\
\cmidrule{2-7}
 & \multirow{2}{*}{OWL}      & Magnitude & 0.0778\textsubscript{±0.0157} & 0.3620\textsubscript{±0.0275} & 0.4500\textsubscript{±0.0000} & 0.7227\textsubscript{±0.0226} \\
 &                           & Wanda     & 0.0778\textsubscript{±0.0157} & 0.3451\textsubscript{±0.0086} & 0.4333\textsubscript{±0.0825} & 0.7120\textsubscript{±0.0113} \\
\cmidrule{2-7}
 & \multirow{2}{*}{LayerIF}  & Magnitude & 0.0778\textsubscript{±0.0416} & 0.3636\textsubscript{±0.0149} & 0.3833\textsubscript{±0.0425} & 0.6960\textsubscript{±0.0123} \\
 &                           & Wanda     & 0.0556\textsubscript{±0.0157} & 0.3333\textsubscript{±0.0258} & 0.4417\textsubscript{±0.0514} & 0.6940\textsubscript{±0.0114} \\
\midrule \midrule
\multirow{6}{*}{Qwen3-8B}
 & \multirow{2}{*}{Uniform}  & Magnitude & 0.5889\textsubscript{±0.0314} & 0.4394\textsubscript{±0.0109} & 0.8583\textsubscript{±0.0236} & 0.9087\textsubscript{±0.0025} \\
 &                           & Wanda     & 0.4889\textsubscript{±0.0685} & 0.4545\textsubscript{±0.0041} & 0.8000\textsubscript{±0.0204} & 0.8967\textsubscript{±0.0034} \\
\cmidrule{2-7}
 & \multirow{2}{*}{OWL}      & Magnitude & 0.5667\textsubscript{±0.0544} & 0.4933\textsubscript{±0.0635} & 0.8667\textsubscript{±0.0624} & 0.9187\textsubscript{±0.0075} \\
 &                           & Wanda     & 0.5000\textsubscript{±0.0000} & 0.4798\textsubscript{±0.0635} & 0.7750\textsubscript{±0.0890} & 0.9027\textsubscript{±0.0074} \\
\cmidrule{2-7}
 & \multirow{2}{*}{LayerIF}  & Magnitude & 0.5222\textsubscript{±0.0629} & 0.4579\textsubscript{±0.0145} & 0.8250\textsubscript{±0.0000} & 0.9107\textsubscript{±0.0090} \\
 &                           & Wanda     & 0.4889\textsubscript{±0.0629} & 0.4747\textsubscript{±0.0149} & 0.8417\textsubscript{±0.0236} & 0.9000\textsubscript{±0.0177} \\
\bottomrule
\end{tabular}
\label{tab:10_1024}
\end{table*}

% ── 2048 tokens ──
\begin{table*}[t]
\centering
\small
\caption{Results for s1.1-7B and Qwen3-8B across allocation strategies and pruning approaches (Global Sparsity: 10\%, Max Thinking Tokens: 2048).}
\begin{tabular}{lll|cccc}
\toprule
\textbf{Model} & \textbf{Allocation} & \textbf{Pruning} & \textbf{AIME24} & \textbf{GPQA-Diamond} & \textbf{AMC23} & \textbf{MATH500} \\
\midrule
\multirow{6}{*}{s1.1-7B}
 & \multirow{2}{*}{Uniform}  & Magnitude & 0.0778\textsubscript{±0.0157} & 0.3532\textsubscript{±0.0188} & 0.4667\textsubscript{±0.0624} & 0.7667\textsubscript{±0.0203} \\
 &                           & Wanda     & 0.0998\textsubscript{±0.0275} & 0.3513\textsubscript{±0.0193} & 0.4917\textsubscript{±0.0425} & 0.7600\textsubscript{±0.0098} \\
\cmidrule{2-7}
 & \multirow{2}{*}{OWL}      & Magnitude & 0.1333\textsubscript{±0.0471} & 0.3367\textsubscript{±0.0086} & 0.4667\textsubscript{±0.0312} & 0.7667\textsubscript{±0.0137} \\
 &                           & Wanda     & 0.1222\textsubscript{±0.0416} & 0.3620\textsubscript{±0.0172} & 0.4750\textsubscript{±0.0736} & 0.7553\textsubscript{±0.0009} \\
\cmidrule{2-7}
 & \multirow{2}{*}{LayerIF}  & Magnitude & 0.0556\textsubscript{±0.0157} & 0.3434\textsubscript{±0.0124} & 0.5000\textsubscript{±0.0250} & 0.7620\textsubscript{±0.0107} \\
 &                           & Wanda     & 0.1000\textsubscript{±0.0272} & 0.3552\textsubscript{±0.0133} & 0.4917\textsubscript{±0.0118} & 0.7600\textsubscript{±0.0065} \\
\midrule \midrule
\multirow{6}{*}{Qwen3-8B}
 & \multirow{2}{*}{Uniform}  & Magnitude & 0.5889\textsubscript{±0.0567} & 0.5152\textsubscript{±0.0189} & 0.9083\textsubscript{±0.0514} & 0.9193\textsubscript{±0.0034} \\
 &                           & Wanda     & 0.5111\textsubscript{±0.0314} & 0.5051\textsubscript{±0.0412} & 0.7667\textsubscript{±0.0773} & 0.9093\textsubscript{±0.0057} \\
\cmidrule{2-7}
 & \multirow{2}{*}{OWL}      & Magnitude & 0.5333\textsubscript{±0.0720} & 0.5084\textsubscript{±0.0401} & 0.8833\textsubscript{±0.0118} & 0.9293\textsubscript{±0.0019} \\
 &                           & Wanda     & 0.5667\textsubscript{±0.0720} & 0.4966\textsubscript{±0.0401} & 0.8667\textsubscript{±0.0236} & 0.9167\textsubscript{±0.0066} \\
\cmidrule{2-7}
 & \multirow{2}{*}{LayerIF}  & Magnitude & 0.5778\textsubscript{±0.0416} & 0.4798\textsubscript{±0.0218} & 0.8750\textsubscript{±0.0354} & 0.9260\textsubscript{±0.0082} \\
 &                           & Wanda     & 0.4556\textsubscript{±0.0157} & 0.4983\textsubscript{±0.0208} & 0.7833\textsubscript{±0.0425} & 0.9140\textsubscript{±0.0033} \\
\bottomrule
\end{tabular}
\label{tab:10_2048}
\end{table*}

% ── 4096 tokens ──
\begin{table*}[t]
\centering
\small
\caption{Results for s1.1-7B and Qwen3-8B across allocation strategies and pruning approaches (Global Sparsity: 10\%, Max Thinking Tokens: 4096).}
\begin{tabular}{lll|cccc}
\toprule
\textbf{Model} & \textbf{Allocation} & \textbf{Pruning} & \textbf{AIME24} & \textbf{GPQA-Diamond} & \textbf{AMC23} & \textbf{MATH500} \\
\midrule
\multirow{6}{*}{s1.1-7B}
 & \multirow{2}{*}{Uniform}  & Magnitude & 0.1331\textsubscript{±0.0269} & 0.3532\textsubscript{±0.0166} & 0.5750\textsubscript{±0.0540} & 0.7847\textsubscript{±0.0096} \\
 &                           & Wanda     & 0.1553\textsubscript{±0.0156} & 0.3530\textsubscript{±0.0326} & 0.5750\textsubscript{±0.0354} & 0.7913\textsubscript{±0.0136} \\
\cmidrule{2-7}
 & \multirow{2}{*}{OWL}      & Magnitude & 0.1444\textsubscript{±0.0416} & 0.3569\textsubscript{±0.0167} & 0.5083\textsubscript{±0.0118} & 0.7853\textsubscript{±0.0090} \\
 &                           & Wanda     & 0.1222\textsubscript{±0.0685} & 0.3434\textsubscript{±0.0071} & 0.5250\textsubscript{±0.0000} & 0.7853\textsubscript{±0.0009} \\
\cmidrule{2-7}
 & \multirow{2}{*}{LayerIF}  & Magnitude & 0.1222\textsubscript{±0.0157} & 0.3687\textsubscript{±0.0165} & 0.5167\textsubscript{±0.0514} & 0.7880\textsubscript{±0.0128} \\
 &                           & Wanda     & 0.1333\textsubscript{±0.0272} & 0.3923\textsubscript{±0.0190} & 0.5000\textsubscript{±0.0408} & 0.7880\textsubscript{±0.0059} \\
\midrule \midrule
\multirow{6}{*}{Qwen3-8B}
 & \multirow{2}{*}{Uniform}  & Magnitude & 0.5444\textsubscript{±0.0567} & 0.5337\textsubscript{±0.0104} & 0.8917\textsubscript{±0.0236} & 0.9313\textsubscript{±0.0066} \\
 &                           & Wanda     & 0.5111\textsubscript{±0.0314} & 0.5522\textsubscript{±0.0333} & 0.9000\textsubscript{±0.0204} & 0.9293\textsubscript{±0.0009} \\
\cmidrule{2-7}
 & \multirow{2}{*}{OWL}      & Magnitude & 0.4778\textsubscript{±0.0567} & 0.5286\textsubscript{±0.0370} & 0.8750\textsubscript{±0.0612} & 0.9353\textsubscript{±0.0009} \\
 &                           & Wanda     & 0.4667\textsubscript{±0.0720} & 0.5370\textsubscript{±0.0370} & 0.8667\textsubscript{±0.0118} & 0.9360\textsubscript{±0.0086} \\
\cmidrule{2-7}
 & \multirow{2}{*}{LayerIF}  & Magnitude & 0.4444\textsubscript{±0.0875} & 0.5219\textsubscript{±0.0186} & 0.9000\textsubscript{±0.0408} & 0.9400\textsubscript{±0.0075} \\
 &                           & Wanda     & 0.5000\textsubscript{±0.0272} & 0.5438\textsubscript{±0.0104} & 0.8750\textsubscript{±0.0354} & 0.9293\textsubscript{±0.0019} \\
\bottomrule
\end{tabular}
\label{tab:10_4096}
\end{table*}

% ── 8192 tokens ──
\begin{table*}[t]
\centering
\small
\caption{Results for s1.1-7B and Qwen3-8B across allocation strategies and pruning approaches (Global Sparsity: 10\%, Max Thinking Tokens: 8192).}
\begin{tabular}{lll|cccc}
\toprule
\textbf{Model} & \textbf{Allocation} & \textbf{Pruning} & \textbf{AIME24} & \textbf{GPQA-Diamond} & \textbf{AMC23} & \textbf{MATH500} \\
\midrule
\multirow{6}{*}{s1.1-7B}
 & \multirow{2}{*}{Uniform}  & Magnitude & 0.1440\textsubscript{±0.0311} & 0.3800\textsubscript{±0.0244} & 0.5667\textsubscript{±0.0118} & 0.8133\textsubscript{±0.0025} \\
 &                           & Wanda     & 0.1889\textsubscript{±0.0157} & 0.3900\textsubscript{±0.0196} & 0.5333\textsubscript{±0.0118} & 0.8033\textsubscript{±0.0098} \\
\cmidrule{2-7}
 & \multirow{2}{*}{OWL}      & Magnitude & 0.1667\textsubscript{±0.0272} & 0.3687\textsubscript{±0.0143} & 0.5417\textsubscript{±0.0425} & 0.8080\textsubscript{±0.0075} \\
 &                           & Wanda     & 0.1111\textsubscript{±0.0416} & 0.3872\textsubscript{±0.0063} & 0.5417\textsubscript{±0.0312} & 0.8047\textsubscript{±0.0096} \\
\cmidrule{2-7}
 & \multirow{2}{*}{LayerIF}  & Magnitude & 0.1333\textsubscript{±0.0272} & 0.3653\textsubscript{±0.0104} & 0.5667\textsubscript{±0.0118} & 0.8080\textsubscript{±0.0043} \\
 &                           & Wanda     & 0.1667\textsubscript{±0.0272} & 0.4192\textsubscript{±0.0051} & 0.5750\textsubscript{±0.0408} & 0.8000\textsubscript{±0.0071} \\
\midrule \midrule
\multirow{6}{*}{Qwen3-8B}
 & \multirow{2}{*}{Uniform}  & Magnitude & 0.6556\textsubscript{±0.0416} & 0.5673\textsubscript{±0.0370} & 0.9083\textsubscript{±0.0312} & 0.9540\textsubscript{±0.0098} \\
 &                           & Wanda     & 0.6000\textsubscript{±0.0471} & 0.5808\textsubscript{±0.0214} & 0.8833\textsubscript{±0.0514} & 0.9540\textsubscript{±0.0049} \\
\cmidrule{2-7}
 & \multirow{2}{*}{OWL}      & Magnitude & 0.5778\textsubscript{±0.0416} & 0.5993\textsubscript{±0.0126} & 0.9000\textsubscript{±0.0204} & 0.9513\textsubscript{±0.0041} \\
 &                           & Wanda     & 0.5778\textsubscript{±0.0416} & 0.5993\textsubscript{±0.0126} & 0.8667\textsubscript{±0.0118} & 0.9580\textsubscript{±0.0075} \\
\cmidrule{2-7}
 & \multirow{2}{*}{LayerIF}  & Magnitude & 0.6000\textsubscript{±0.0471} & 0.5808\textsubscript{±0.0071} & 0.8917\textsubscript{±0.0118} & 0.9573\textsubscript{±0.0041} \\
 &                           & Wanda     & 0.6111\textsubscript{±0.0314} & 0.6077\textsubscript{±0.0242} & 0.9000\textsubscript{±0.0204} & 0.9520\textsubscript{±0.0020} \\
\bottomrule
\end{tabular}
\label{tab:10_8192}
\end{table*}

\begin{table*}[t]
\centering
\small
\caption{Qualitative outputs on AIME24 for Unpruned vs.\ Wanda-Uniform 20\% pruned model.}
\label{tab:cot_comparison}
\begin{tabular}{m{3cm}|m{5.5cm}|m{5.5cm}}
\toprule
\textbf{Query} & \textbf{Unpruned} & \textbf{Wanda-Uniform 20\%} \\
\midrule
Let $p$ be the least prime number for which there exists a positive integer $n$ such that $n^{4}+1$ is divisible by $p^{2}$. Find the least positive integer $m$ such that $m^{4}+1$ is divisible by $p^{2}$. & \textbf{Chain of Thought:} Okay, so I need to find the smallest prime number p where there's a positive integer n such that $p^{2}$ divides $n^{4} +1$. Then, once I find that $p$, I need to find the smallest positive integer $m$ such that $p^{2}$  divides $m^4 + 1$. Hmm. Let me start by understanding the problem step by step. First, let's think about primes $p$ where $p^{2}$  divides $n^{4} +1$ for some $n$. That means $n^4 = -1 mod p^2$. So, $n^4 = -1$ mod $p^2$. Let me recall that if a number a is congruent to $-1$ modulo $p^2$, then raising it to the power of 4 gives us a result that's congruent to $1 \pmod{p^2}$ because $(-1)^4 = 1$. Wait, but $n^4 \equiv -1 \pmod{p^2}$. So, squar..............is odd, then p is $1 mod 2$. Suppose $p = 1 mod 4$. Then $p - 1$ is divisible by $4$. Let me check: If $p = 1 mod 4$, then $p - 1$ is divisible by $4$, so $k = (p - 1)/2$ is divisible by $2$. Therefore, $k*p$ is divisible by $2 * p$, which is 2*(something odd). Therefore, to have k*p divisible by 4, since k is divisible by 2 (as$ p = 1 mod 4$), p is odd, so k*p is divisible by 2*odd, but we need it to be divisible by 4. Therefore, if k is divisible by 2, then k*p is divisible by $2*2 = 4$? Wait, k is $(p - 1)/2$. If $p = 1$\newline \textcolor{red}{\textbf{Answer: 155 (Incorrect)}} & \textbf{Chain of Thought:} Okay, so I need to find the least prime number $p$ such that there's a positive integer $n$ where $p^2$ divides $n^4 + 1$. Then, once I find that $p$, I need to find the smallest positive integer $m$ such that $p^2$ divides $m^4 + 1$. Hmm. Let me start by understanding the problem step by step. First, let's think about primes $p$ where $p^2$ divides $n^4 + 1$. That means $n^4 \equiv -1 \pmod{p^2}$. So, we need to find primes $p$ for which $-1$ is a quartic residue modulo $p$.............Alternatively, compute $n^4 \bmod 11$: $n=0$: $0$, $n=1$: $1$, $n=2$: $16 \bmod 11 = 5$, $n=3$: $81 \bmod 11$. $11*7 = 77$, so $81-77 = 4$. $3^4 = 4 \bmod 11$, $n=4$: $4^2 = 16 \bmod 11 = 5$; $4^4 = 5^2 = 25 \bmod 11 = 3$, $n=5$: $5^2 = 25 \bmod 11 = 3$; $5^4 = 3^2 = 9 \bmod 11$, $n=6$: $6^2 = 36 \bmod 11 = 3$; $6^4 = 9 \bmod 11$, $n=7$: $7^2 = 49 \bmod 11 = 5$, then $7^4 = 5^2 = 25 \bmod 11 = 3$, $n=8$: $8^2 = 64 \bmod 11 = 9$, so $8^4 = 9^2 = 81 \bmod 11 = 4$, $n=9$: $9^2 = 81 \bmod 11 = 4$, then $9^4 = 4^2 = 16 \bmod 11 = 5$, $n=10$: $10^2 = 100 \bmod 11 = 1$ (since $11*9 = 99$, $100-99 = 1$). \newline \textcolor{teal}{\textbf{Answer: 110 (Correct)}} \\
\midrule\midrule
Let $\mathcal{B}$ be the set of rectangular boxes with surface area $54$ and volume $23$. Let $r$ be the radius of the smallest sphere that can contain each of the rectangular boxes that are elements of $\mathcal{B}$. The value of $r^2$ can be written as $\frac{p}{q}$, where $p$ and $q$ are relatively prime positive integers. Find $p+q$. & 
\textbf{Chain of Thought:} Okay, so I need to find the smallest sphere that can contain any rectangular box with surface area \(54\) and volume \(23\). Then, find \(r^2\) where \(r\) is the radius of that sphere, and express it as a fraction \(p/q\) with \(p\) and \(q\) coprime, then find \(p+q\). Alright, let's start breaking this down. First, let me recall that for a rectangular box with dimensions \(x, y, z\), the surface area is \(2(xy + yz + zx)\) and the volume is \(xyz\). So, given that the surface area is \(54\) and the volume is \(23\), I can write: $2(xy + yz + zx) = 54 \quad \Rightarrow \quad xy + yz + zx = 27,$ ........, , xy).

Therefore, the equations are: 

$2x = \lambda (y + z) + \mu yz$,  
$2y = \lambda (x + z) + \mu xz$,  
$2z = \lambda (x + y) + \mu xy$.

Additionally, we have the constraints: 

$xy + yz + zx = 27$,  
$xyz = 23$.

So, four equations with variables $x, y, z, \lambda, \mu$. That seems complicated. However, perhaps there is some symmetry.

Alternatively, maybe we can assume some symmetry. For example, suppose two variables are equal. Let's assume $x = y$. Then, as before, we can solve. But earlier, we found solutions. However, maybe that is the case where the
and $xyz = 23.$
So, the problem is about rectangles \(x, y, z\) with these.
\newline \textcolor{red}{\textbf{Answer: 137 (Incorrect)}}& 
\textbf{Chain of Thought:} Okay, so I need to find the smallest sphere that can contain any rectangular box with surface area $54$ and volume $23$. Then, find $r^2$ where $r$ is the radius of that sphere, and express it as a fraction $p/q$ with $p$ and $q$ coprime, then find $p+q$. Alright, let's start.

First, let me recall some formulas. For a rectangular box with length, width, and height $x, y, z$, the surface area is $2(xy + yz + xz) = 54$. The volume is $xyz = 23$. The problem is about finding the smallest sphere that can contain such a box. The sphere's radius will depend on the box's dimensions. ..........= $-1 + 2\sqrt{6} \approx -1 + 4.898 \approx 3.898$, which is positive.

Therefore, if $x = y = 2$, then $z = 23/(2^2) = 23/4 = 5.75$. Alternatively, if $x = y \approx 3.898$, then $z = 23/x^2$. Let me compute $x^2$ exactly: $x = -1 + 2\sqrt{6}$. Therefore, $x^2 = (-1 + 2\sqrt{6})^2 = 1 - 4\sqrt{6} + 24 = 25 - 4\sqrt{6} \approx 25 - 9.798 \approx 15.202$. Then $z = 23 / (25 - 4\sqrt{6})$. Let me rationalize that denominator:

Multiply numerator and denominator by $(25 + 4\sqrt{6})$:

$z = \frac{23 (25 + 4\sqrt{6})}{(25 - 4\sqrt{6})(25 + 4\sqrt{6})} = \frac{23 (25 + 4\sqrt{6})}{625 - (4\sqrt{6})^2}$. \newline \textcolor{teal}{\textbf{Answer: 721 (Correct)}}\\

\bottomrule
\end{tabular}
\end{table*}

\begin{table*}[t]
\centering
\small
\caption{Qualitative outputs on AMC23 for Unpruned vs.\ Wanda-Uniform 20\% pruned model.}
\label{tab:cot_comparison_amc}
\begin{tabular}{m{3cm}|m{5.5cm}|m{5.5cm}}
\toprule
\textbf{Query} & \textbf{Unpruned} & \textbf{Wanda-Uniform 20\%} \\
\midrule

If $A$ and $B$ are vertices of a polyhedron, define the distance $d(A,B)$ to be the minimum number of edges of the polyhedron one must traverse in order to connect $A$ and $B$. For example, if $\overline{AB}$ is an edge of the polyhedron, then $d(A, B) = 1$, but if $\overline{AC}$ and $\overline{CB}$ are edges and $\overline{AB}$ is not an edge, then $d(A, B) = 2$. Let $Q$, $R$, and $S$ be randomly chosen distinct vertices of a regular icosahedron (regular polyhedron made up of 20 equilateral triangles). Find the probability that $d(Q, R) > d(R, S)$. The final answer can be written in the form $\frac{m}{n}$, where $m$ and $n$ are relatively prime positive integers. What is $m+n$? & \textbf{Chain of Thought:} Okay, so I need to find the maximum value of the imaginary part of a complex number z that satisfies the equation $|1 + z + z^2| = 4$. The answer should be in the form $\frac{\sqrt(m)}{ n}$, where m and n are coprime, and then find m + n. Alright, let's start by recalling some complex number properties. First, let me denote z as a complex number. Let me write $z = x + yi$, where x and y are real numbers. Then, I need to compute $|1 + z + z^2|$ and set it equal to 4, then find the maximum value of y. But before diving into algebra, maybe there's a smarter way. Sometimes with modulus equations, it's helpful to consi
..........$y^4 + (2x^2+ 2x -1)y^2 - 16 = 0$. Hmm, this seems complicated. Maybe there's a better approach. Let me think. Alternatively, maybe consider substituting $y^2$ from the equation or something else. Alternatively, think of this as a quadratic equation in $y^2$? Let me check. Let me rearrange the equation: $y^4 + (2x^2 + 2x -1)y^2 + (x^4 + 2x^3 + 3x^1 + 2x +1 - 16) = 0$. So, if I let $y^2 = t$, then:$ t^2 + (2x^2 + 2x -1)t + (x^4 + 2x^3 + 3x^2 + 2x +1 - 16) = 0$. Let me compute the constant term: $x^4 + 2x^2 + 3x^2$
\newline \textcolor{red}{\textbf{Answer: 31 (Incorrect)}}
& \textbf{Chain of Thought:} Okay, so I need to find the maximum value of the imaginary part of a complex number z that satisfies the equation $|1 + z + z^2| = 4$. The answer should be in the form $\frac{\sqrt{m}}{n}$, where m and n are coprime, and then find $m + n$. Alright, let's start by understanding the problem. First, z is a complex number. Let me denote $z = x + yi$, where x and y are real numbers. The imaginary part of z is y, so we need to maximize y given that$|1 + z + z^2| = 4$. Let me write down the equation $|1 + z + z^2| = 4$. Let me compute $1 + z + z^2$ first. Since $z = x + yi$, then $z^2 = (x + yi)^2 = x^2 - y^2 + 2xyi$. Therefore, 1 + ............... y, try to parametrize z. Let me think. Let me consider that $1 + z + z^2 = 4e^{i\theta}$, where $\theta$ is some angle. Then, solving for z. But solving quadratic equations with complex coefficients might not be straightforward. Alternatively, maybe consider that z is a root of the equation $z^2 + z + 1 = 4e^{i\theta}$, so$z^2 + z + (1 - 4e^{i\theta}) = \theta$. Then, solving for $z: z = [-1 \pm \sqrt{(1 - 4(1 - 4e^{i\theta}))}]/2 = [-1 \pm \sqrt{(1 - 4 + 16e^{i\theta})}]/2 = [-1 \pm \sqrt{(-3 + 16e^{i\theta})}]/2$. But this seems complicated as well. The square root
\newline \textcolor{teal}{\textbf{Answer: 29 (Correct)}}\\
\midrule\midrule
In the state of Coinland, coins have values $6,10,$ and $15$ cents. Suppose $x$ is the value in cents of the most expensive item in Coinland that cannot be purchased using these coins with exact change. What is the sum of the digits of $x?$ & \textbf{Chain of Thought:} Okay, so I need to figure out the largest amount of money that can't be made with coins of 6, 10, and 15 cents. Then, once I find that number, called x, I have to find the sum of its digits. Alright, let me start by recalling some math concepts. First, I remember something called the Frobenius number, which is the largest monetary amount that cannot be obtained using any combination of coins of specified denominations. But wait, the Frobenius number is only straightforward to calculate when there are two coin denominations. For more than two, there isn't a general formula, right? So, since we .............  but we need the largest one. So, if we can make all numbers from 45 upwards, then 44 would be the answer. But since we have 15, maybe 44 can be made with 15? Wait, 44. Let me check. Let's see, can we make 44 with 6, 10, 15? Let's try. First, try using 15 once: 44 - 15 = 29. Can we make 29 with 6 and 10? 29 is odd. 6 and 10 are even. So no. If we use 15 twice: 44 - 30 = 14. Can we make 14 with 6 and 10? 10 + 6* (something)? 14 -10 = 4. Not divisible by 6. 14 - 6 = 8. Not divisible by 10. So no
\newline \textcolor{red}{\textbf{Answer: 8 (Incorrect)}}
&\textbf{Chain of Thought:} Okay, so I need to figure out the largest amount of money that can't be formed using coins of 6, 10, and 15 cents. Then, once I find that number, called x, I have to find the sum of its digits. Alright, let me start by recalling some concepts about the Frobenius number. Wait, the Frobenius number is the largest number that can't be expressed as a combination of given denominations, right? But I remember that the Frobenius number is only straightforward to calculate when there are two denominations. For more than two, there isn't a general formula, so I guess I need another approach here. The ............. formed. 5*4 = 20 can be formed as 10+10. So the only number that cannot be formed with 10 and 15 is 5. So in terms of numbers that can be formed using 10 and 15, they are all multiples of 5 starting from 10. So if we have 5, it can't be formed. But 5 is not a multiple of 6, 10, or 15. So if we add 6 to the equation, perhaps we can form numbers that aren't multiples of 5? Wait, but the problem is we can use 6, 10, 15. So maybe the idea is that if we can use 6 to adjust the amounts. Let me think.
\newline \textcolor{teal}{\textbf{Answer: 11 (Correct)}}
\\

\bottomrule
\end{tabular}
\end{table*}

\begin{table*}[t]
\centering
\small
\caption{Qualitative outputs on GPQA-Diamond for Unpruned vs.\ Wanda-Uniform 20\% pruned model.}
\label{tab:cot_comparison_gpqa}
\begin{tabular}{m{3cm}|m{5.5cm}|m{5.5cm}}
\toprule
\textbf{Query} & \textbf{Unpruned} & \textbf{Wanda-Uniform 20\%} \\
\midrule
ChIP-seq on a PFA-fixed sample with an antibody to the IKAROS transcription factor in human B cells followed by next-generation sequencing and standard quality control, alignment and peak-calling steps produced ChIP peaks that disappeared when PFA+DSG fixation was used. Where are we most likely to find such disappearing peaks? 
\newline A. At random locations in the genome \newline B. At active promoters and enhancers \newline C. At repeats \newline D. In the introns of large genes & \textbf{Chain of Thought} Okay, so I need to figure out where the ChIP peaks that disappear when using PFA+DSG fixation compared to just PFA fixation would be located. Let me start by recalling what I know about ChIP-seq and fixation methods. First, ChIP-seq is used to identify the binding sites of proteins, like transcription factors, across the genome. The process involves fixing the proteins to DNA using a cross-linking agent, then immunoprecipitating the protein-DNA complexes, and sequencing the DNA. PFA (paraformaldehyde) is a common fixative used in ChIP experiments. It's a mild cross-linker that primarily cros .......... ucture, and if the fixation with PFA+DSG alters the chromatin structure or protein-DNA interactions in these regions, the peaks would disappear. </think> When analyzing the disappearance of ChIP-seq peaks in samples fixed with **PFA+DSG** compared to **PFA alone**, we focus on the **mechanism of fixation and its impact on protein-DNA interactions**. Here's a breakdown of the reasoning: ---  **Key Considerations** 1. **Fixation and Protein-DNA Interactions**: - **PFA (paraformaldehyde)** \newline \textcolor{red}{\textbf{Answer: A (Incorrect)}}&  \textbf{Chain of Thought} Okay, so I need to figure out where the disappearing ChIP peaks are most likely located when using PFA+DSG fixation instead of just PFA. Let me start by recalling what I know about ChIP-seq and fixation methods. First, ChIP-seq is used to determine where a particular protein binds to DNA. The process involves crosslinking the protein to DNA, shearing the DNA, and then immunoprecipitating the protein-DNA complexes. The DNA is then sequenced to find the binding sites. Now, fixation with PFA (paraformaldehyde) is a common method. But sometimes, other fixatives are used. The question mentions DS .......................... ing some issue with the crosslinking in certain regions. Alternatively, think about the fact that some fixatives like formaldehyde are better at preserving protein-DNA interactions, while others may not. If the DSG fixation is not as effective, then the peaks that require strong crosslinking would be lost. If those peaks are in regions where the protein is bound more tightly, maybe in active promoters or enhancers. Alternatively, if the DSG fixation is causing more shearing in certain regions \newline \textcolor{teal}{\textbf{Answer: B (Correct)}}\\
\midrule\midrule
A benzene ring has a carboxylic acid a carbaldehyde and a cyano group all meta to one another. Ortho to the carboxylic acid are a hydroxyl and a dimethyl amino and para to the carboxylic acid is a methoxy group. The methoxy and the alcohol are also both ortho to the nitrile. What is the correct IUPAC name for this molecule? \newline A. 2-hydroxy-3-cyano-4-methoxy-5-formyl-6-(dimethylamino)benzoic acid \newline B. 3-cyano-6-(dimethylamino)-5-formyl-2-hydroxy-4-methoxybenzoic acid \newline C. 2-(dimethylamino)-3-formyl-4-methoxy-5-cyano-6-hydroxybenzoic acid \newline D. 5-cyano-2-(dimethylamino)-3-formyl-6-hydroxy-4-methoxybenzoic acid &  \textbf{Chain of Thought:} Okay, so I need to figure out the correct IUPAC name for this complex benzene ring compound. Let me start by breaking down the information given. The molecule has a benzene ring with four substituents: a carboxylic acid, a carbaldehyde (which is formyl group), a cyano group, and a methoxy group. Additionally, there's a hydroxyl group and a dimethylamino group. The positions of these groups relative to each other are specified as meta, ortho, and para. Let me start by recalling that in IUPAC nomenclature for aromatic compounds, the substituents are numbered in such a way that the lowest possible numbers are assigned, considering the order of priority. The priority order for substituents is usually based on their functional groups, with carboxylic acids being higher priority than others. Wait, but in the case of benzoic acids, the carbox .................  4, 5, 6. Suppose the carboxylic acid is at position 1. Then, to have another group meta to it, it must be at 3. Then, the third group would need to be meta to both. If the second group is at 3 (meta to 1), then the third group must be meta to both 1 and 3. The positions that are meta to 1 are 3 and 5. The positions that are meta to 3 are 1 and 5. Therefore, the only common position is 5. So if the third group is at 5, then it is meta to both 1 and 3. Therefore, the three groups are at positions
\newline \textcolor{red}{\textbf{Answer: C (Incorrect)}}
& \textbf{Chain of Thought:}  Okay, so I need to figure out the correct IUPAC name for this complex benzene ring molecule. Let me start by breaking down the information given. First, the molecule has a benzene ring with four different functional groups: a carboxylic acid, a carbaldehyde (which is an aldehyde group), a cyano group (CN), and a hydroxyl group (OH). Additionally, there are two more substituents: a dimethylamino group (which is an N,N-dimethylamino group) and a methoxy group (OCH3). The problem states that these groups are arranged in specific positions relative to each other. Let me parse that step by step. The first part says: "A benzene ring has a carboxylic acid, a carbaldehyde, and a cyano group all meta to one another." So, the carboxylic acid, aldehyde, and cyano group are all meta to each other. ................ ption B. Wait, but let me check the options again. Option B is: 3-cyano-6-(dimethylamino)-5-formyl-2-hydroxy-4-methoxybenzoic acid Yes. That seems to match my analysis. But let me check if the order of substituents is correct. In IUPAC nomenclature, substituents are listed in alphabetical order. Let me check the substituents again. The substituents are: - cyano (CN) - dimethylamino (N(CH3)2) - formyl (CHO) - hydroxyl (OH) - methoxy (OCH3) So, in alphabetical order, the order should   \newline \textcolor{teal}{\textbf{Answer: B (Correct)}}
\\

\bottomrule
\end{tabular}
\end{table*}

\begin{table*}[t]
\centering
\small
\caption{Qualitative outputs on MATH500 for Unpruned vs.\ Wanda-Uniform 20\% pruned model. }
\label{tab:cot_comparison_math}
\begin{tabular}{m{3cm}|m{5.5cm}|m{5.5cm}}
\toprule
\textbf{Query} & \textbf{Unpruned} & \textbf{Wanda-Uniform 20\%} \\
\midrule
The expression $2\cdot 3 \cdot 4\cdot 5+1$ is equal to 121, since multiplication is carried out before addition. However, we can obtain values other than 121 for this expression if we are allowed to change it by inserting parentheses. For example, we can obtain 144 by writing 
$(2\cdot (3\cdot 4)) \cdot (5+1) = 144$In total, how many values can be obtained from the expression $2\cdot 3\cdot 4 \cdot 5 + 1$ by inserting parentheses? (Note that rearranging terms is not allowed, only inserting parentheses). & \textbf{Chain of Thought:} Okay, so I need to figure out how many different values can be obtained from the expression 2·3·4·5 + 1 by inserting parentheses. The original value is 121 because multiplication is done before addition. But if I can add parentheses, maybe I can get different results. The example given is (2·(3·4))·(5+1) = 144. So, I need to explore all possible ways of inserting parentheses here and see how many unique results we can get. First, let me recall that inserting parentheses changes the order of operations. Since multiplication is associative and commutative, but addition is only associative, so inserting parentheses might change the order in which operations are performed, especially since we have both multiplication and addition here. Wait, but in the original expression, the .................. grouping: (2 * ( (3 + 4) * 5 )) + 1. Let me compute: 3 +4=7, 7*5=35, 2*35=70, 70 +1=71. Which we had earlier. How about (2 * 3) * (4 + 5) + 1. That's 6 * 9 +1 = 54 +1 = 55. Which we had before. How about ((2 + 3) * 4 * 5) +1? Wait, but the original expression doesn't have a + between 2 and 3. Wait, but can we insert parentheses to do (2 + 3) * 4 * 5 +1? Wait, but the original expression is 2*3*4*5 +1. So inserting parentheses to do (2 + 3) would require changing the operations? Wait, no. Wait,  \newline \textcolor{red}{\textbf{Answer: 8 (Incorrect)}} 
&  \textbf{Chain of Thought:}  Okay, so I need to figure out how many different values can be obtained from the expression 2·3·4·5 + 1 by inserting parentheses. The original value is 121 because multiplication is done before addition. But if I can add parentheses, maybe I can get different results. The example given is 144 by doing (2·(3·4))·(5+1). Let me try to understand this. First, the original expression is 2·3·4·5 + 1. Since multiplication has higher precedence than addition, it's evaluated as (2·3·4·5) + 1 = 120 + 1 = 121. But if I can insert parentheses, I can group operations differently. The problem says that rearranging terms isn't allowed, only inserting parentheses. So I can't change the order of the numbers, only how the operations are grouped. Let me think about the structure of the expression. The original ............ : between the product of 2,3,4,5 and 1. So the addition is only between the entire product and 1. So, to have the addition happen earlier, we need to have parentheses that include the addition operator. But the addition is between the product and 1, so how can we group it differently? Wait, for example, maybe inserting parentheses between the multiplications and the addition. For example, 2·3·(4·(5 + 1)) which is same as before. Or 2·(3·4·(5 + 1)). All these groupings lead to same value. So the \newline \textcolor{teal}{\textbf{Answer: 4 (Correct)}} \\
\midrule\midrule
Let $p(x)$ be a polynomial of degree 5 such that $p(n) = \frac{n}{n^2 - 1}$for $n = 2,$ 3, 4, $\dots,$ 7. Find $p(8).$
& \textbf{Chain of Thought:}  Okay, so I need to find p(8) where p(x) is a degree 5 polynomial that satisfies $p(n) = n/(n^2 - 1)$ for n = 2, 3, 4, 5, 6, 7. Hmm. Let me think. First, since p(x) is a degree 5 polynomial and we have 6 points (from n=2 to n=7), that should uniquely determine the polynomial, right? Because a polynomial of degree n is uniquely determined by n+1 points. So here, degree 5, 6 points, so unique. But how do I find p(8)? But maybe there's a smarter way than just finding the polynomial directly? Let me recall some techniques. Maybe interpolation? But interpolation for degree 5 might be tedious. Maybe there's a trick here with the given function $n/(n^2 - 1)$. Let ..........  function $f(x) = x/(x^2 - 1)$. For $x = 8, f(8) = 8/(64 - 1) = 8/63 \approx 0.12698$. But p(8) is $721/9 \approx 80.111$. That seems way larger. But p(x) is a degree 5 polynomial passing through those points. So maybe it's possible? But maybe there's a mistake in the calculation? Because the value seems very different from f(8). Let me check q(8) again. Wait, $q(x) = (x^2 - 1)p(x) - x.$ So, if I think about x=8, then: q(8) = (64 - 1)p(8) - 8 = 63p(8) - 8. But we \newline \textcolor{red}{\textbf{Answer: $\frac{721}{9}$  (Incorrect)}}  
& \textbf{Chain of Thought:} Okay, so I need to find p(8) where p(x) is a degree 5 polynomial that satisfies $p(n) = n/(n^2 - 1)$ for n = 2, 3, 4, 5, 6, 7. Hmm. Let me think. First, since p(x) is a degree 5 polynomial and we have 6 points (from n=2 to n=7), that should uniquely determine the polynomial, right? Because a polynomial of degree n is uniquely determined by n+1 points. So here, degree 5, 6 points, so unique. But how do I find p(8)? But maybe there's a smarter way than constructing the entire polynomial? Maybe using some kind of interpolation or something? Let me recall that when you have a polynomia  ............  Therefore, $Q(x) = (x - 2)(x - 3)...(x - 7)(x - a)$, where a is some constant. Therefore, Q(x) is degree 7. But then, how can I find k? Wait, but I need another condition. Maybe we can find Q(x) by evaluating at another point? For example, we can plug in x = 1? Let me try that. Let me compute $Q(1) = (1^2 - 1)p(1) - 1 = 0 * p(1) - 1 = -1.$ On the other hand, $Q(1) = (1 - 2)(1 - 3)...(1 - 7)(1 - a) = (-1)(-2)(-3)(-4)(-5)(-6)(1 - a)$. Let me compute that product. First, compute the product (1 - 2)(1  \newline \textcolor{teal}{\textbf{Answer: $\frac{3}{56}$ (Correct)}}
\\
\bottomrule
\end{tabular}
\end{table*}

% \begin{table*}[t]
% \centering
% \small
% \caption{LLM as judge score using GPT-4o on the Chain of Thought reasoning between Unpruned, Wanda-10\%, and Wanda-20\% sparsity on AIME24 for 2048 and 8192 thinking tokens. Results indicate that, although the differences are modest, the Wanda20 model consistently achieves higher judge scores than the unpruned baseline.}
% \begin{tabular}{c|ccc}
% \toprule
% \textbf{Max\_Token} & \textbf{Unpruned} & \textbf{Wanda 10\%} & \textbf{Wanda 20\%} \\
% \midrule
% 2048 & 2.500 & 2.500 & \textbf{2.630} \\
% 8192 & 2.230 & 2.170 & \textbf{2.300} \\
% \bottomrule
% \end{tabular}
% \label{tab:llm_judge}
% \end{table*}

\begin{table*}[t]
\centering
\small
\caption{LLM as judge score using GPT-4o on the Chain of Thought reasoning between Unpruned and Wanda-20\% sparsity on AIME24 for 2048 and 8192 thinking tokens. Results indicate that, although the differences are modest, the Wanda-20\% model consistently achieves higher judge scores than the unpruned baseline.}
\begin{tabular}{c|cc}
\toprule
\textbf{Max\_Token} & \textbf{Unpruned} & \textbf{Wanda 20\%} \\
\midrule
2048 & 2.500 &  \textbf{2.630} \\
8192 & 2.230 &  \textbf{2.300} \\
\bottomrule
\end{tabular}
\label{tab:llm_judge}
\end{table*}

\begin{table*}[t]
\centering
\small
\caption{Results for Qwen3-8B on AIME24 and GPQA-Diamond comparing exact parameter count pruned using unstructured (magnitude and wanda) and structured pruning (ShortGPT). Here, 1 Layer and 2 Layer indicate 1 or 2 Layers removed using ShortGPT. For Qwen3-8B, removing one layer amounts to 2.7\% and two layers amount to 5.5\% removal of total parameters.}
\begin{tabular}{ll|cccccccc}
\toprule
\textbf{Dataset} & \textbf{Token} & \textbf{Unpruned} & \textbf{1 Layer} & \textbf{2 Layer} & \textbf{Mag 2.7\%} & \textbf{Mag 5.5\%} & \textbf{Wanda 2.7\%} & \textbf{Wanda 5.5\%} \\
\midrule
\multirow{5}{*}{AIME}
 & 512  & \textbf{0.633\textsubscript{±0.072}} & 0.378\textsubscript{±0.057} & 0.100\textsubscript{±0.027} & 0.610\textsubscript{±0.016} & \textbf{0.633\textsubscript{±0.000}} & 0.533\textsubscript{±0.100} & 0.550\textsubscript{±0.050} \\
 & 1024 & 0.489\textsubscript{±0.057} & 0.289\textsubscript{±0.042} & 0.067\textsubscript{±0.027} & 0.500\textsubscript{±0.100} & 0.516\textsubscript{±0.050} & 0.466\textsubscript{±0.033} & \textbf{0.580\textsubscript{±0.080}} \\
 & 2048 & 0.433\textsubscript{±0.000} & 0.333\textsubscript{±0.094} & 0.078\textsubscript{±0.031} & 0.433\textsubscript{±0.033} & 0.450\textsubscript{±0.017} & \textbf{0.500\textsubscript{±0.000}} & 0.430\textsubscript{±0.030} \\
 & 4096 & 0.433\textsubscript{±0.047} & 0.289\textsubscript{±0.031} & 0.100\textsubscript{±0.027} & 0.450\textsubscript{±0.050} & 0.430\textsubscript{±0.000} & \textbf{0.517\textsubscript{±0.016}} & 0.450\textsubscript{±0.016} \\
 & 8192 & 0.578\textsubscript{±0.042} & 0.411\textsubscript{±0.042} & 0.133\textsubscript{±0.000} & 0.560\textsubscript{±0.060} & 0.580\textsubscript{±0.016} & 0.580\textsubscript{±0.110} & \textbf{0.610\textsubscript{±0.016}} \\
\midrule
\multirow{5}{*}{GPQA}
 & 512  & 0.446\textsubscript{±0.020} & 0.399\textsubscript{±0.021} & 0.384\textsubscript{±0.030} & \textbf{0.455\textsubscript{±0.015}} & 0.442\textsubscript{±0.023} & 0.452\textsubscript{±0.013} & 0.450\textsubscript{±0.005} \\
 & 1024 & 0.466\textsubscript{±0.010} & 0.423\textsubscript{±0.013} & 0.397\textsubscript{±0.021} & 0.470\textsubscript{±0.040} & \textbf{0.475\textsubscript{±0.010}} & 0.470\textsubscript{±0.005} & 0.450\textsubscript{±0.010} \\
 & 2048 & 0.493\textsubscript{±0.037} & 0.455\textsubscript{±0.011} & 0.406\textsubscript{±0.019} & 0.513\textsubscript{±0.013} & 0.505\textsubscript{±0.010} & \textbf{0.518\textsubscript{±0.033}} & 0.508\textsubscript{±0.013} \\
 & 4096 & 0.493\textsubscript{±0.037} & 0.498\textsubscript{±0.013} & 0.409\textsubscript{±0.025} & \textbf{0.561\textsubscript{±0.025}} & \textbf{0.561\textsubscript{±0.020}} & 0.538\textsubscript{±0.033} & 0.556\textsubscript{±0.005} \\
 & 8192 & 0.588\textsubscript{±0.019} & 0.540\textsubscript{±0.030} & 0.421\textsubscript{±0.023} & 0.596\textsubscript{±0.000} & 0.606\textsubscript{±0.005} & 0.581\textsubscript{±0.010} & \textbf{0.619\textsubscript{±0.008}} \\
\bottomrule
\end{tabular}
\label{tab:qwen3_results_equal}
\end{table*}

\begin{table*}[t]
\centering
\small
\caption{Results for s1.1-7B on AIME24 and GPQA-Diamond comparing exact parameter count pruned using unstructured (magnitude and wanda) and structured pruning (ShortGPT). Here, 1 Layer and 2 Layer indicate 1 or 2 Layers removed using ShortGPT. For s1.1-7B, removing one layer amounts to 3.5\% and two layers amount to 7\% removal of total parameters.}
\begin{tabular}{ll|cccccccc}
\toprule
\textbf{Dataset} & \textbf{Token} & \textbf{Unpruned} & \textbf{1 Layer} & \textbf{2 Layer} & \textbf{Mag 3.5\%} & \textbf{Mag 7\%} & \textbf{Wanda 3.5\%} & \textbf{Wanda 7\%} \\
\midrule
\multirow{5}{*}{AIME}
 & 512  & 0.078\textsubscript{±0.016} & 0.011\textsubscript{±0.016} & 0.011\textsubscript{±0.016} & 0.067\textsubscript{±0.000} & \textbf{0.117\textsubscript{±0.017}} & 0.067\textsubscript{±0.000} & 0.017\textsubscript{±0.017} \\
 & 1024 & 0.078\textsubscript{±0.032} & 0.033\textsubscript{±0.027} & 0.003\textsubscript{±0.005} & \textbf{0.100\textsubscript{±0.033}} & 0.050\textsubscript{±0.050} & 0.083\textsubscript{±0.017} & 0.083\textsubscript{±0.050} \\
 & 2048 & \textbf{0.122\textsubscript{±0.016}} & 0.011\textsubscript{±0.016} & 0.003\textsubscript{±0.005} & 0.083\textsubscript{±0.017} & 0.100\textsubscript{±0.000} & 0.067\textsubscript{±0.033} & 0.083\textsubscript{±0.017} \\
 & 4096 & 0.133\textsubscript{±0.000} & 0.044\textsubscript{±0.042} & 0.003\textsubscript{±0.005} & 0.133\textsubscript{±0.000} & 0.083\textsubscript{±0.017} & \textbf{0.150\textsubscript{±0.050}} & \textbf{0.150\textsubscript{±0.017}} \\
 & 8192 & 0.167\textsubscript{±0.027} & 0.044\textsubscript{±0.016} & 0.011\textsubscript{±0.016} & 0.183\textsubscript{±0.050} & \textbf{0.217\textsubscript{±0.083}} & 0.117\textsubscript{±0.017} & 0.183\textsubscript{±0.017} \\
\midrule
\multirow{5}{*}{GPQA}
 & 512  & \textbf{0.345\textsubscript{±0.027}} & 0.315\textsubscript{±0.009} & 0.320\textsubscript{±0.009} & 0.331\textsubscript{±0.008} & 0.323\textsubscript{±0.030} & 0.331\textsubscript{±0.023} & 0.326\textsubscript{±0.013} \\
 & 1024 & 0.338\textsubscript{±0.008} & 0.313\textsubscript{±0.031} & 0.288\textsubscript{±0.036} & 0.290\textsubscript{±0.013} & 0.328\textsubscript{±0.025} & 0.328\textsubscript{±0.005} & \textbf{0.354\textsubscript{±0.005}} \\
 & 2048 & \textbf{0.353\textsubscript{±0.014}} & 0.342\textsubscript{±0.025} & 0.294\textsubscript{±0.024} & 0.333\textsubscript{±0.020} & 0.336\textsubscript{±0.018} & 0.338\textsubscript{±0.015} & 0.343\textsubscript{±0.005} \\
 & 4096 & 0.362\textsubscript{±0.022} & 0.343\textsubscript{±0.019} & 0.292\textsubscript{±0.026} & \textbf{0.381\textsubscript{±0.018}} & 0.346\textsubscript{±0.003} & 0.381\textsubscript{±0.003} & \textbf{0.394\textsubscript{±0.005}} \\
 & 8192 & \textbf{0.391\textsubscript{±0.034}} & 0.330\textsubscript{±0.019} & 0.301\textsubscript{±0.035} & 0.381\textsubscript{±0.033} & 0.364\textsubscript{±0.025} & 0.389\textsubscript{±0.025} & 0.369\textsubscript{±0.020} \\
\bottomrule
\end{tabular}
\label{tab:s1_results_equal}
\end{table*}

\begin{table*}[t]
\centering
\small
\caption{Additional results for Llama-3.1-8B (DeepSeek-R1-Distill) on AIME24 and GPQA-Diamond pruned using unstructured (magnitude and wanda) and structured pruning (ShortGPT). Here, 1 Layer and 2 Layer indicate 1 or 2 Layers removed using ShortGPT.}
\begin{tabular}{ll|cccccccc}
\toprule
\textbf{Dataset} & \textbf{Token} & \textbf{Unpruned} & \textbf{1 Layer} & \textbf{2 Layer} & \textbf{Mag 10\%} & \textbf{Mag 20\%} & \textbf{Wanda 10\%} & \textbf{Wanda 20\%} \\
\midrule
\multirow{5}{*}{AIME}
 & 512  & \textbf{0.380\textsubscript{±0.016}} & 0.300\textsubscript{±0.067} & 0.183\textsubscript{±0.083} & \textbf{0.380\textsubscript{±0.500}} & 0.350\textsubscript{±0.160} & 0.360\textsubscript{±0.030} & 0.310\textsubscript{±0.830} \\
 & 1024 & \textbf{0.410\textsubscript{±0.016}} & 0.367\textsubscript{±0.033} & 0.100\textsubscript{±0.067} & \textbf{0.410\textsubscript{±0.010}} & 0.360\textsubscript{±0.280} & 0.310\textsubscript{±0.050} & 0.330\textsubscript{±0.000} \\
 & 2048 & \textbf{0.400\textsubscript{±0.130}} & 0.167\textsubscript{±0.033} & 0.200\textsubscript{±0.033} & 0.360\textsubscript{±0.030} & 0.280\textsubscript{±0.050} & 0.310\textsubscript{±0.167} & 0.300\textsubscript{±0.100} \\
 & 4096 & \textbf{0.433\textsubscript{±0.067}} & 0.350\textsubscript{±0.017} & 0.300\textsubscript{±0.000} & 0.410\textsubscript{±0.050} & 0.380\textsubscript{±0.110} & 0.430\textsubscript{±0.000} & 0.360\textsubscript{±0.330} \\
 & 8192 & \textbf{0.460\textsubscript{±0.060}} & 0.233\textsubscript{±0.033} & 0.250\textsubscript{±0.083} & 0.430\textsubscript{±0.000} & 0.360\textsubscript{±0.060} & 0.410\textsubscript{±0.016} & 0.410\textsubscript{±0.083} \\
\midrule
\multirow{5}{*}{GPQA}
 & 512  & 0.350\textsubscript{±0.035} & 0.342\textsubscript{±0.018} & 0.333\textsubscript{±0.010} & \textbf{0.380\textsubscript{±0.002}} & 0.360\textsubscript{±0.035} & \textbf{0.380\textsubscript{±0.035}} & 0.370\textsubscript{±0.003} \\
 & 1024 & \textbf{0.380\textsubscript{±0.050}} & 0.369\textsubscript{±0.015} & 0.376\textsubscript{±0.043} & 0.370\textsubscript{±0.023} & 0.370\textsubscript{±0.005} & \textbf{0.380\textsubscript{±0.013}} & \textbf{0.380\textsubscript{±0.040}} \\
 & 2048 & 0.390\textsubscript{±0.010} & 0.376\textsubscript{±0.008} & 0.376\textsubscript{±0.008} & \textbf{0.400\textsubscript{±0.020}} & 0.370\textsubscript{±0.008} & \textbf{0.400\textsubscript{±0.003}} & 0.390\textsubscript{±0.018} \\
 & 4096 & 0.400\textsubscript{±0.005} & 0.366\textsubscript{±0.053} & 0.338\textsubscript{±0.015} & 0.400\textsubscript{±0.008} & 0.400\textsubscript{±0.008} & 0.410\textsubscript{±0.008} & \textbf{0.420\textsubscript{±0.051}} \\
 & 8192 & \textbf{0.470\textsubscript{±0.020}} & 0.323\textsubscript{±0.141} & 0.328\textsubscript{±0.066} & 0.440\textsubscript{±0.025} & 0.430\textsubscript{±0.023} & 0.450\textsubscript{±0.046} & 0.410\textsubscript{±0.028} \\
\bottomrule
\end{tabular}
\label{tab:llama_results}
\end{table*}

\end{document}